\documentclass[10pt,twocolumn,letterpaper]{article}

\usepackage{cvpr}              

\definecolor{cvprblue}{rgb}{0.21,0.49,0.74}
\usepackage[pagebackref,breaklinks,colorlinks,allcolors=cvprblue]{hyperref}

\def\paperID{*****} 
\def\confName{CVPR}
\def\confYear{2025}

\title{6D Pose Estimation on Spoons and Hands}

\author{Kevin Tan, Fan Yang, Yuhao Chen\\
University of Waterloo\\
{\tt\small \{kx3tan, f25yang, yuhao.chen1\}@uwaterloo.ca}
}

\begin{document}
\maketitle
\begin{abstract}
	Accurate dietary monitoring is essential for promoting healthier eating habits. A key area of research is how people interact and consume food using utensils and hands. By tracking their position and orientation, it is possible to estimate the volume of food being consumed, or monitor eating behaviours, highly useful insights into nutritional intake that can be more reliable than popular methods such as self-reporting. Hence, this paper implements a system that analyzes stationary video feed of people eating, using 6D pose estimation to track hand and spoon movements to capture spatial position and orientation. In doing so, we examine the performance of two state-of-the-art (SOTA) video object segmentation (VOS) models, both quantitatively and qualitatively, and identify main sources of error within the system.
\end{abstract}    
\section{Introduction}
\label{sec:intro}

6D Pose Estimation refers to predicting the spatial position and orientation (pitch, yaw, roll) within an environment. Obtaining accurate estimates for eating utensils like spoons and hands can play a significant role in nutritional and dietary analysis, such as being able to then estimate food intake or monitor eating behaviours. In such cases, the non-intrusive nature of vision-based pose estimation approaches is especially desirable.

This paper introduces an approach to use 6D pose estimation for tracking hands and spoons when eating, enabling a detailed analysis of various actions involved in the consumption of food. This technique is able to overcome common limitations of vision settings such as occlusion, depth ambiguity, and motion blur, by providing robust tracking even with complex movements, making it well-suited for realistic settings where users interact dynamically with a variety of foods.

We then perform a comparative analysis of two SOTA VOS models, and evaluate this system's accuracy in realistic eating scenarios. Ultimately we highlight such a system's potential applications in dietary monitoring research.
\section{Related Works}
\label{sec:related works}

Pose estimation, particularly in dynamic hand-object interactions is a growing area of interest with applications in dietary monitoring, robotics, and augmented reality. This section aims to highlight key methods and models related to 6D pose estimation to track objects like utensils and hands.

Accurately estimating the pose of hand-object interactions has seen many advancements. For example, Wang et al. presents a method that roughly estimates the hand and object poses separately, then combines and refines the results using a graph convolution network and mutual attention layer \cite{wang2022interactinghandobjectposeestimation}. To get the rough pose estimates, they represent each 3D vertex of a hand mesh as quantized pixel coordinates and depth, using a ResNet-50 encoder to extract image features \cite{wang2022interactinghandobjectposeestimation}. 

In addition, Millerdurai et al. present \textit{Ev2Hands} which reconstructs the 3D pose of hands, including the shape, pose, and position, from a monocular event camera stream \cite{millerdurai20233dposeestimationinteracting}. Then it feeds the point-cloud representation into an attention-based neural network \cite{millerdurai20233dposeestimationinteracting}. This method is powerful in its ability to work well in low light or high-speed movement \cite{millerdurai20233dposeestimationinteracting}.

Aboukhadra et al. presents \textit{SurgeoNet} which tackles 7D pose estimation of surgical instruments, including hand-utensil interactions \cite{aboukhadra2024surgeonetrealtime3dpose} which is very related to pose estimation of spoons in a food consumption context. It uses virtual reality glasses to obtain a sequence of RGB images \cite{aboukhadra2024surgeonetrealtime3dpose} and feeds it to a pipeline involving object detection using \textit{Yolov8} \cite{Jocher_Ultralytics_YOLO_2023}, tracking and temporal smoothing of keypoints with \textit{ByteTrack} \cite{zhang2022bytetrackmultiobjecttrackingassociating}, and 7D pose estimation using a Transformer network \cite{aboukhadra2024surgeonetrealtime3dpose}.

Finally, PoseCNN has seen much success in performing 6D pose estimation of known objects especially in cluttered environments where occlusions are common \cite{xiang2018posecnnconvolutionalneuralnetwork}. It makes use of a Convolutional Neural Network to estimate an object's 3D translation and 3D rotation \cite{xiang2018posecnnconvolutionalneuralnetwork}. 

\section{Methodology}
\label{sec:methodology}

\begin{figure*}[th]
    \centering
    \includegraphics[width=\linewidth]{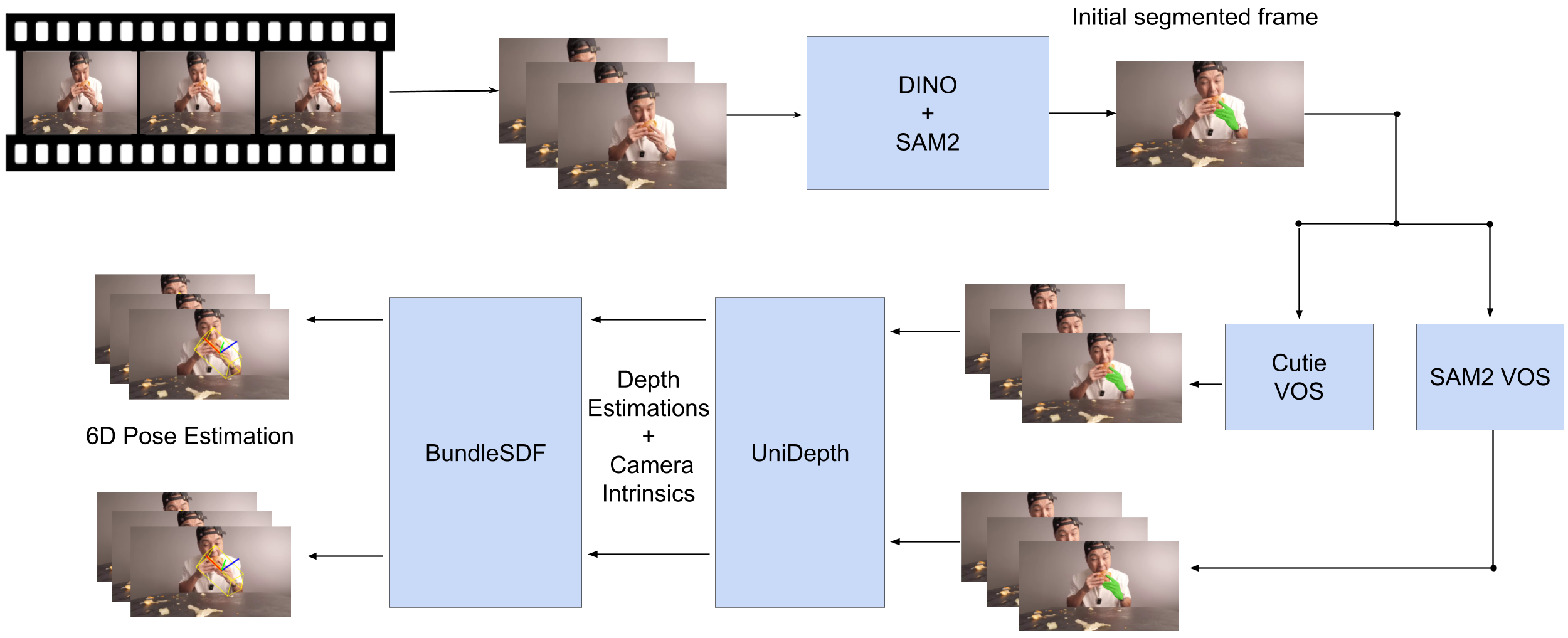}
    \caption{Summary of pose estimation pipeline. Frames are extracted from videos and passed into Grounding DINO and SAM2 for zero-shot segmentation of spoons and hands. The initial segmented frame is passed into either Cutie or SAM2 for video object segmentation and the resulting 6D pose estimation is evaluated.}
    \label{fig:fig_block_diagram}
\end{figure*}

There are three main steps in our video processing pipeline as depicted in \cref{fig:fig_block_diagram}: Zero-shot object detection and segmentation, depth estimation, and 6D pose estimation. 

\subsection{Zero-shot Detection and Segmentation}
Incorporating zero-shot segmentation offers several advantages for tracking and segmenting objects. Spoons and hands come in different shapes and sizes and so it is crucial to select models that can generalize well across various contexts. One approach would be to use a zero-shot detection model such as Grounding DINO \cite{liu2023grounding} to produce bounding boxes as input to a segmentation model such as SAM2 \cite{ravi2024sam2}, to segment each individual frame for a spoon or hand. 

However, exclusively using Grounding DINO and SAM2 to segment objects throughout a video is not only extremely slow, but it also fails to draw on the temporal relationship between frames and thus can produce inconsistent segmentations \cite{sharma2024atefoodportionestimation}. To address these issues, we employ VOS. Specifically, we compare the performance of Cutie \cite{cheng2023putting} and SAM2's video predictor. To obtain the initial segmentation, we still use Grounding DINO and SAM2. 

\begin{figure*}
    \captionsetup[subfigure]{labelformat=empty, position=top, labelfont=bf}
    \centering
    \subfloat[RGB Image]{\frame{\includegraphics[width=.195\linewidth]{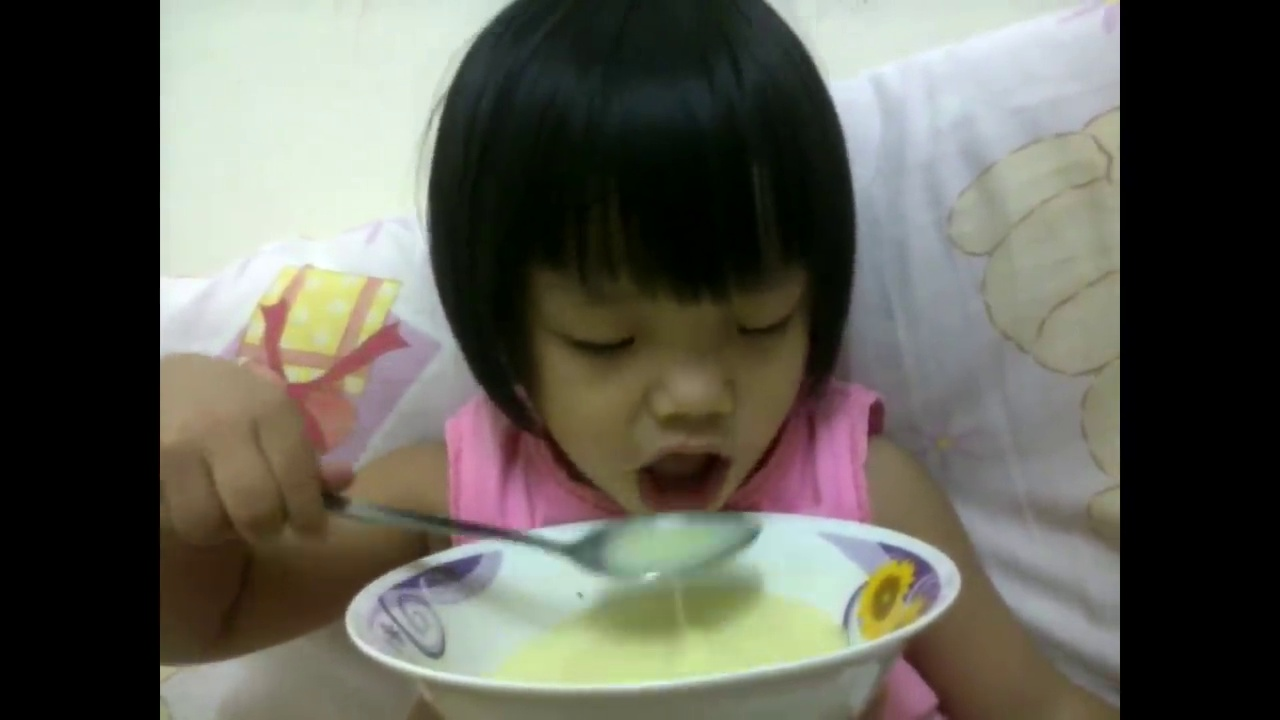}}}
    \subfloat[Cutie Segmentation]{\frame{\includegraphics[width=.195\linewidth]{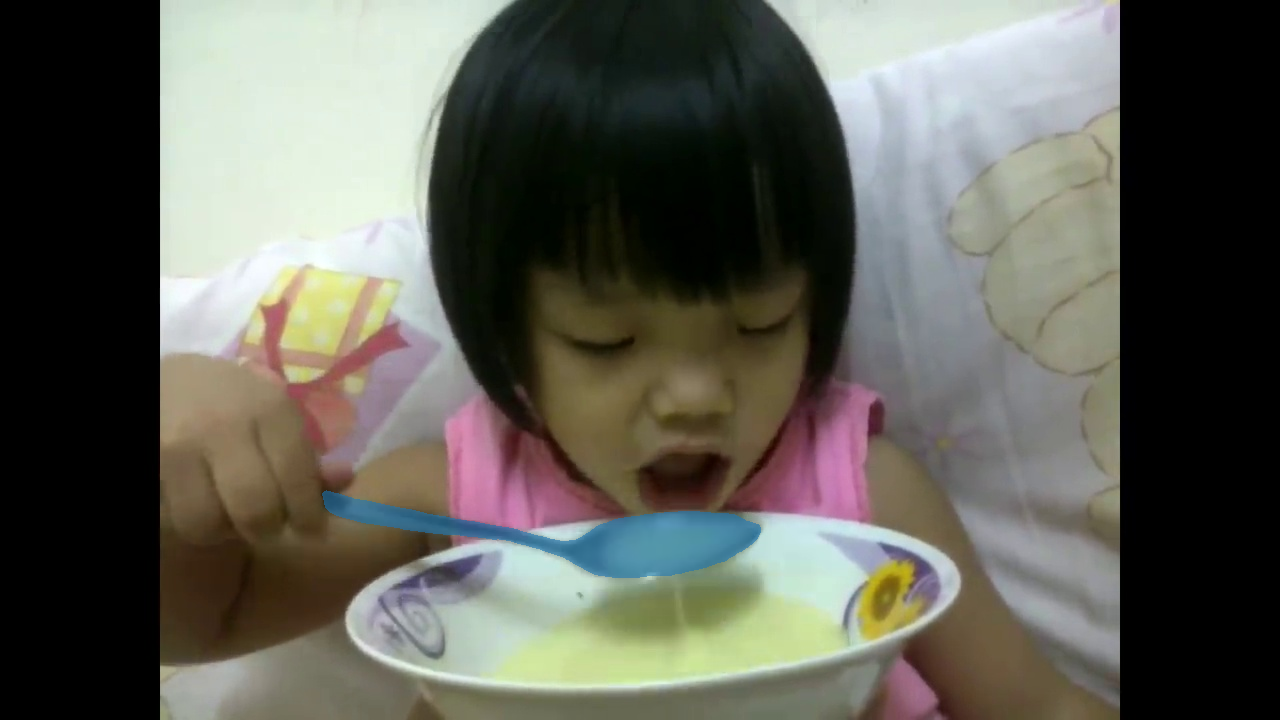}}}
    \subfloat[SAM2 Segmentation]
    {\frame{\includegraphics[width=.195\linewidth]{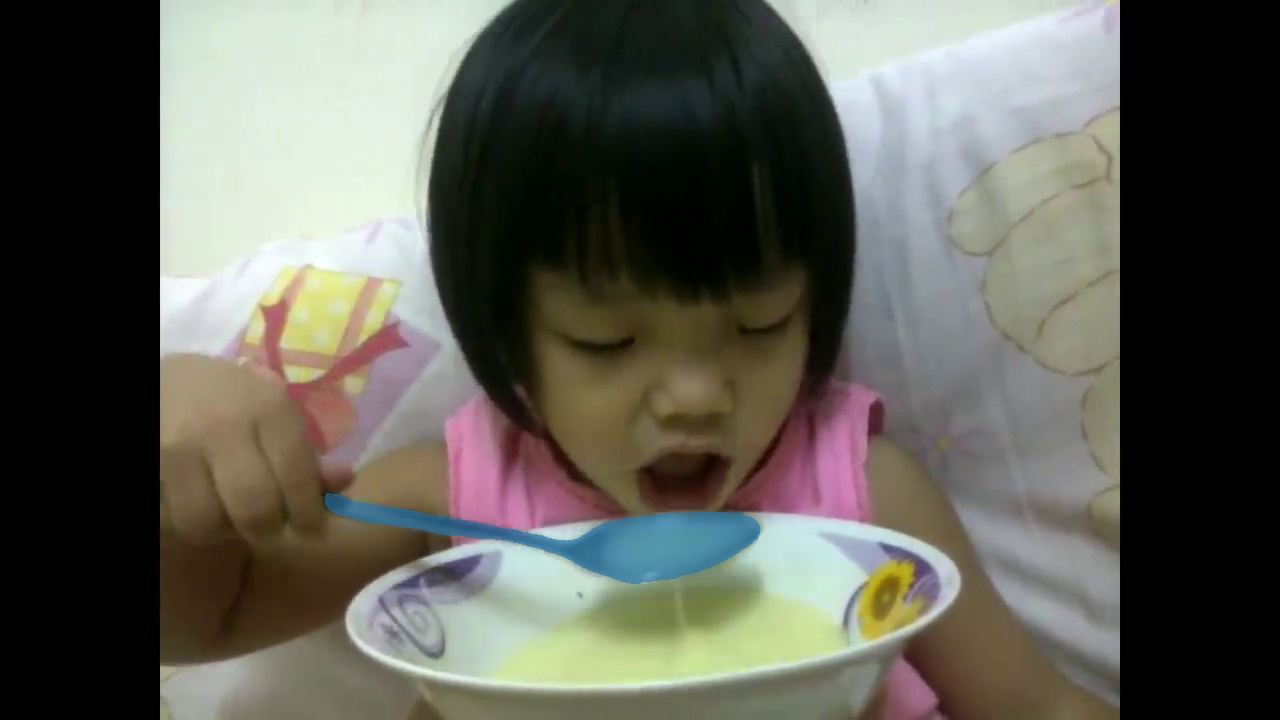}}}
    \subfloat[Cutie Pose Estimation]
    {\frame{\includegraphics[width=.195\linewidth]{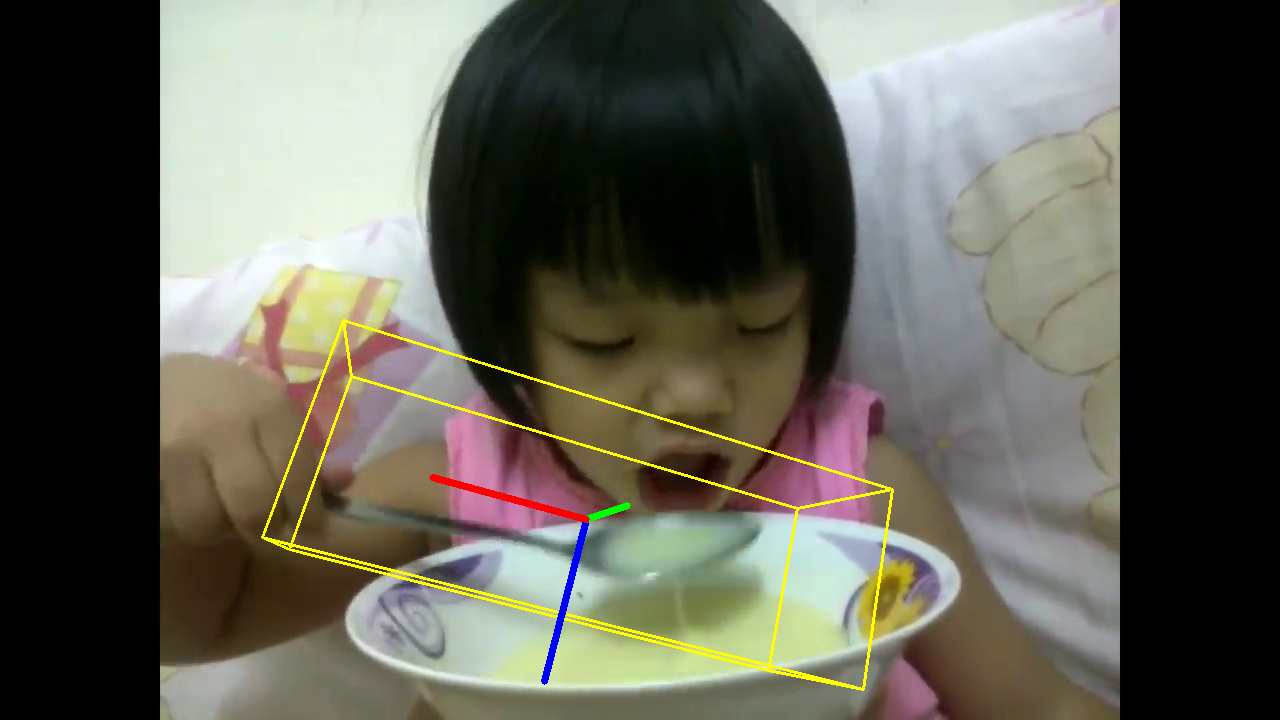}}}
    \subfloat[SAM2 Pose Estimation]{\frame{\includegraphics[width=.195\linewidth]{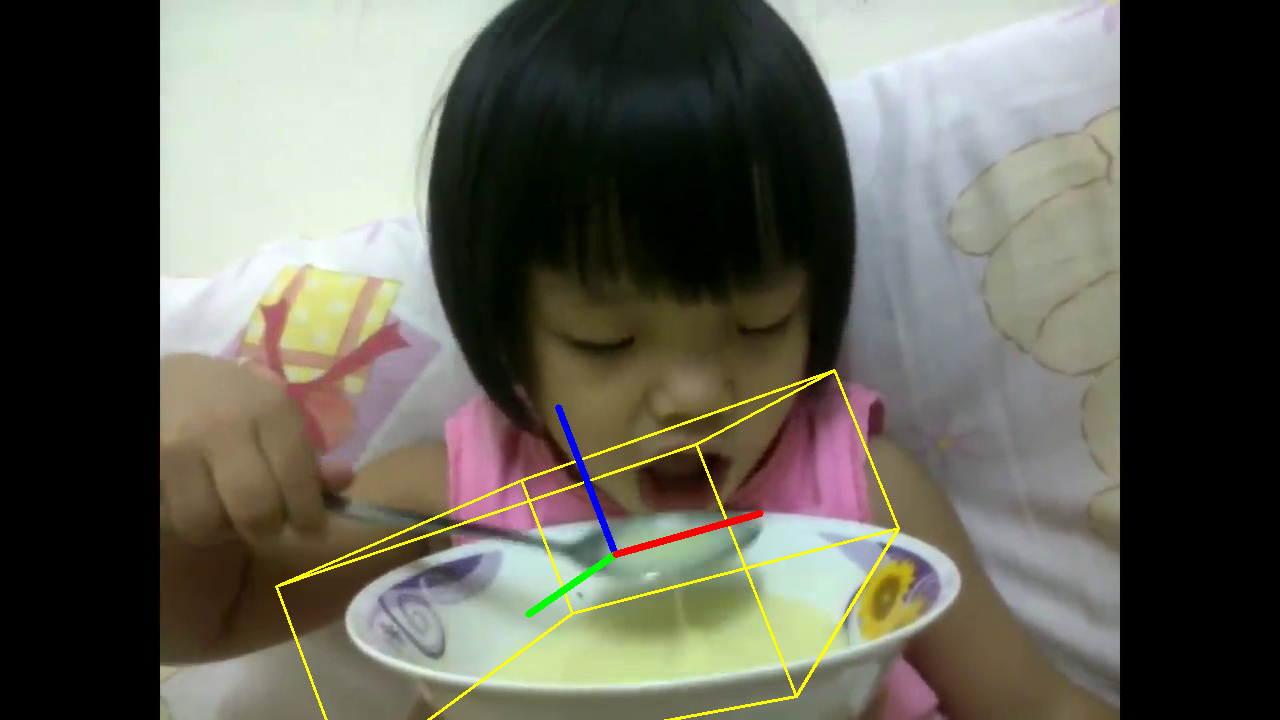}}}
    \vspace{0.1em}
    \subfloat[]{\frame{\includegraphics[width=.195\linewidth]{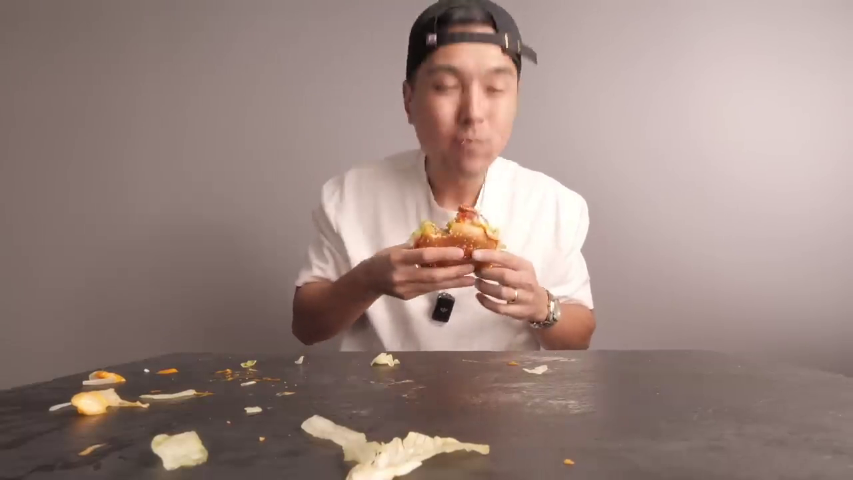}}}
    \subfloat[]{\frame{\includegraphics[width=.195\linewidth]{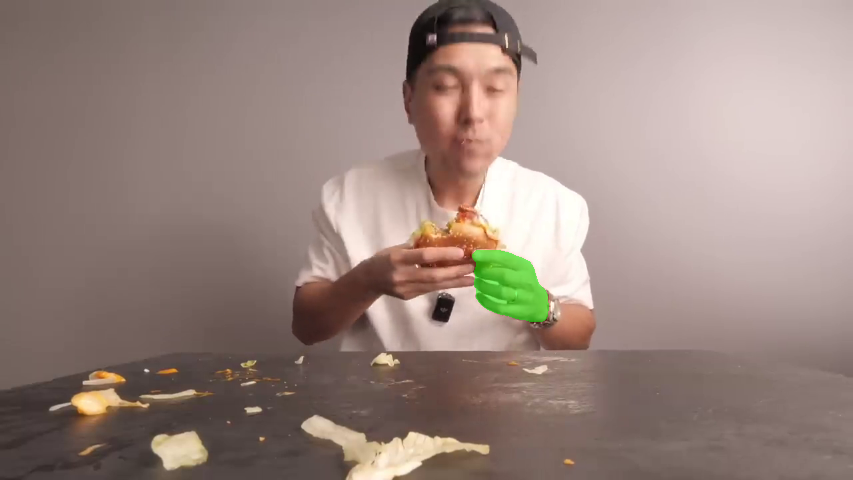}}}
    \subfloat[]{\frame{\includegraphics[width=.195\linewidth]{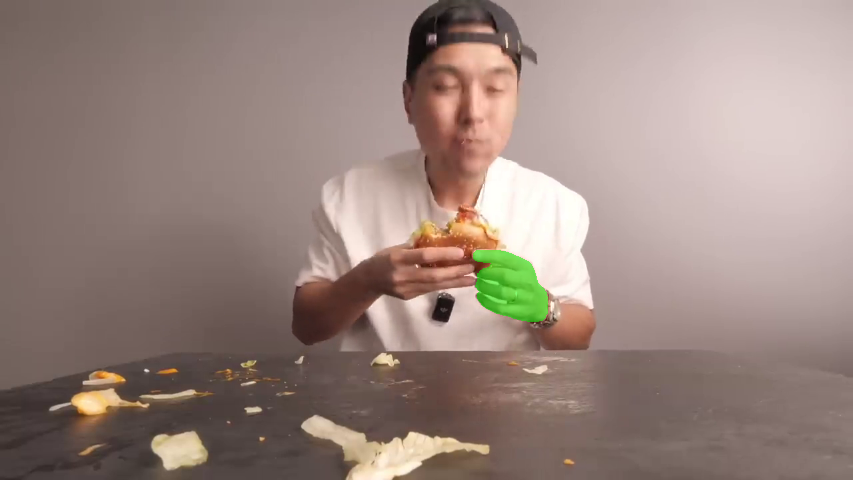}}}
    \subfloat[]{\frame{\includegraphics[width=.195\linewidth]{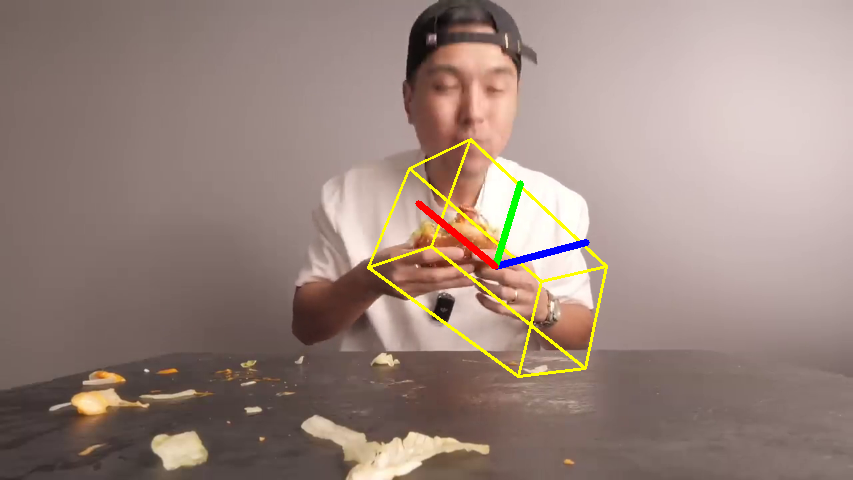}}}
    \subfloat[]
    {\frame{\includegraphics[width=.195\linewidth]{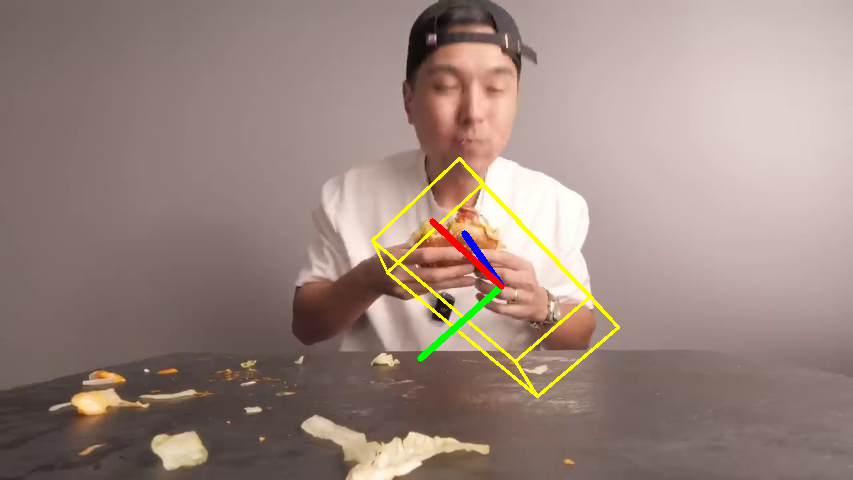}}}
  \caption{Example of segmentation and pose estimation results using the Cutie and SAM2 models. First row displays results for a spoon. Second row displays results for the hand on the right.} 
    \label{fig:process}
\end{figure*}

With this segmentation, we are also able to filter for frames where there is a person eating with their hands or spoon, removing frames of food preparation, talking, or other introductory frames of no interest. The frame filtration involves finding faces with hands, spoon, and food. This filters out most of the irrelevant frames where people are not eating. Refer to \cref{fig:process} to see the result. The prompts used are "Spoon" and "Hand".

\subsection{Depth Estimation} 

Prior to performing 6D Pose Estimation, depth estimation is needed to aid 3D perception. For this, we use UniDepth \cite{piccinelli2024unidepth}, which again provides powerful generalization capabilities without the need for additional context information. By applying UniDepth on each video frame, we obtain depth estimates and the source camera's intrinsic parameters.
\subsection{6D Pose Estimation}
The final stage of the pipeline is 6D pose estimation, This is handled by BundleSDF \cite{bundlesdfwen2023}, which enables 6-DoF tracking of unknown objects when supplied monocular RGB frames, segmentation masks, depth estimates, as well as camera intrinsics. BundleSDF was chosen for its ability to provide robust tracking through large pose changes and occlusions, with the only assumption necessary is that the object is segmented in the first frame.

\subsection{Dataset}

Our dataset consists of two videos of real-world eating scenarios found on Youtube, with each video featuring eating with either hands or a spoon. For time efficiency, a subset of all frames is taken. In particular, the objects of interest in both videos are often occluded or exhibit motion blur.

\section{Results}
\label{sec:results}

We investigate the quality of segmentation from using Cutie and SAM2 (using SAM2-Hiera-Large model), as well as the 6D pose estimates. Our primary focus lies in assessing the accuracy of each model and identifying common failure points within the system. 

\subsection{Segmentation Accuracy}

To provide a baseline measure of segmentation quality, manual annotation was performed on all frames in the video sequences to obtain ground truth masks, and calculate a Dice Similarity Coefficient \cite{https://doi.org/10.2307/1932409} (DSC) for each frame. For two bitmasks $A, B$ representing the ground truth and model segmentation, the DSC can be computed by
\begin{equation}
    \frac{2*|A\cap B|}{|A| + |B|}
\end{equation}
Results are shown in \cref{tab:tab1}.
\begin{table}[ht]
    \centering
    \begin{tabular}{l|c|c|c}
        \hline
        Object & Number of Frames & Cutie & SAM2 \\
        \hline
        Hand & 200 & 0.8133 & 0.8255 \\
        Spoon & 387 & 0.8477 & 0.9286 \\
        \bottomrule
    \end{tabular}
    \caption{Mean Dice Similarity Coefficient}
    \label{tab:tab1}
\end{table}

The scores indicate that SAM2 performs stronger than Cutie on average in segmenting both hands and spoons.


\subsection{Segmentation Failures}

\begin{figure}[h]
    \centering Cutie
    \vskip 0.2em
    \begin{tabular}{@{}c@{}c}
        \includegraphics[width=.5\linewidth]{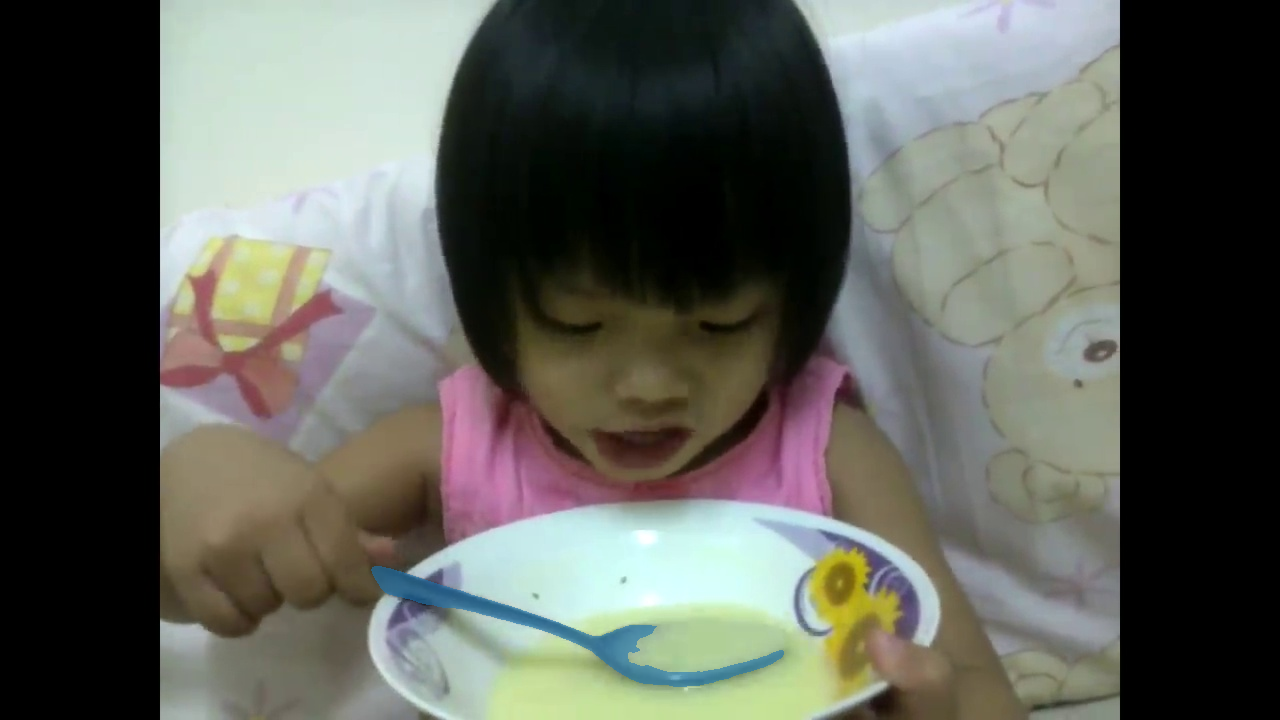} &
        \includegraphics[width=.5\linewidth]{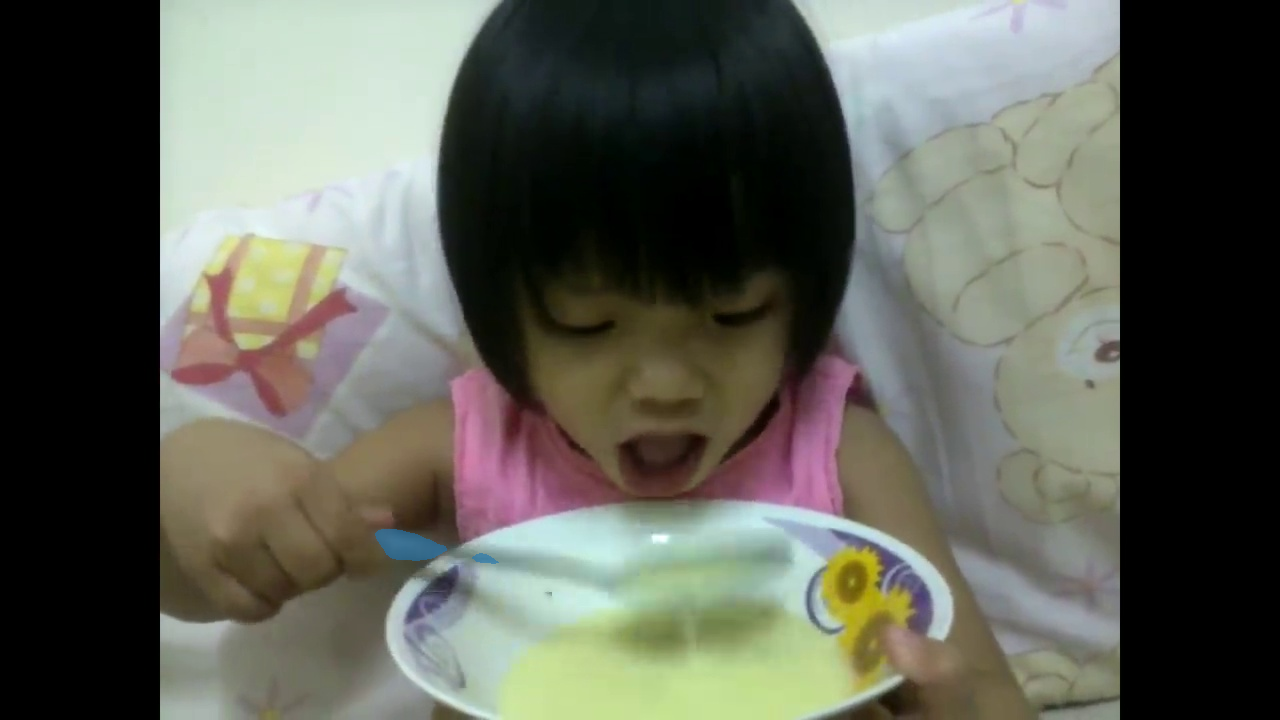} \\
    \end{tabular}
    \vskip 0.5em
    \centering SAM2
    \vskip 0.2em    
    \begin{tabular}{@{}c@{}c}
        \includegraphics[width=.5\linewidth]{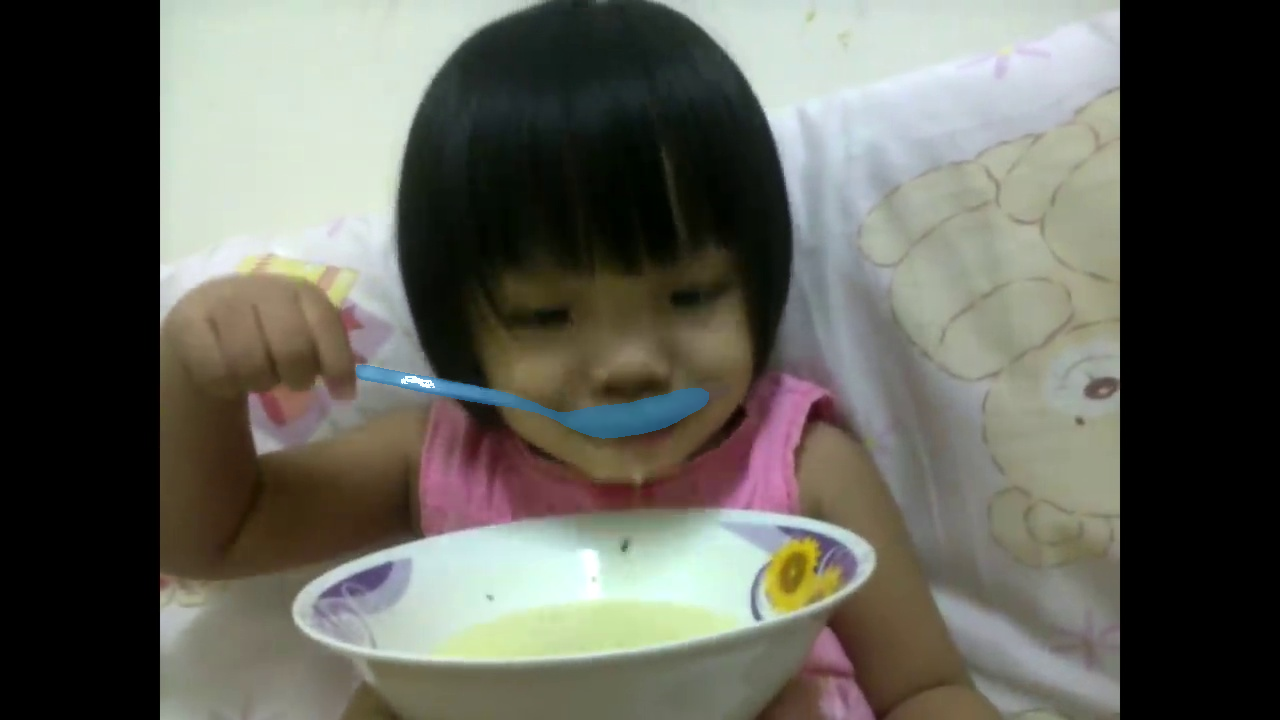} &
        \includegraphics[width=.5\linewidth]{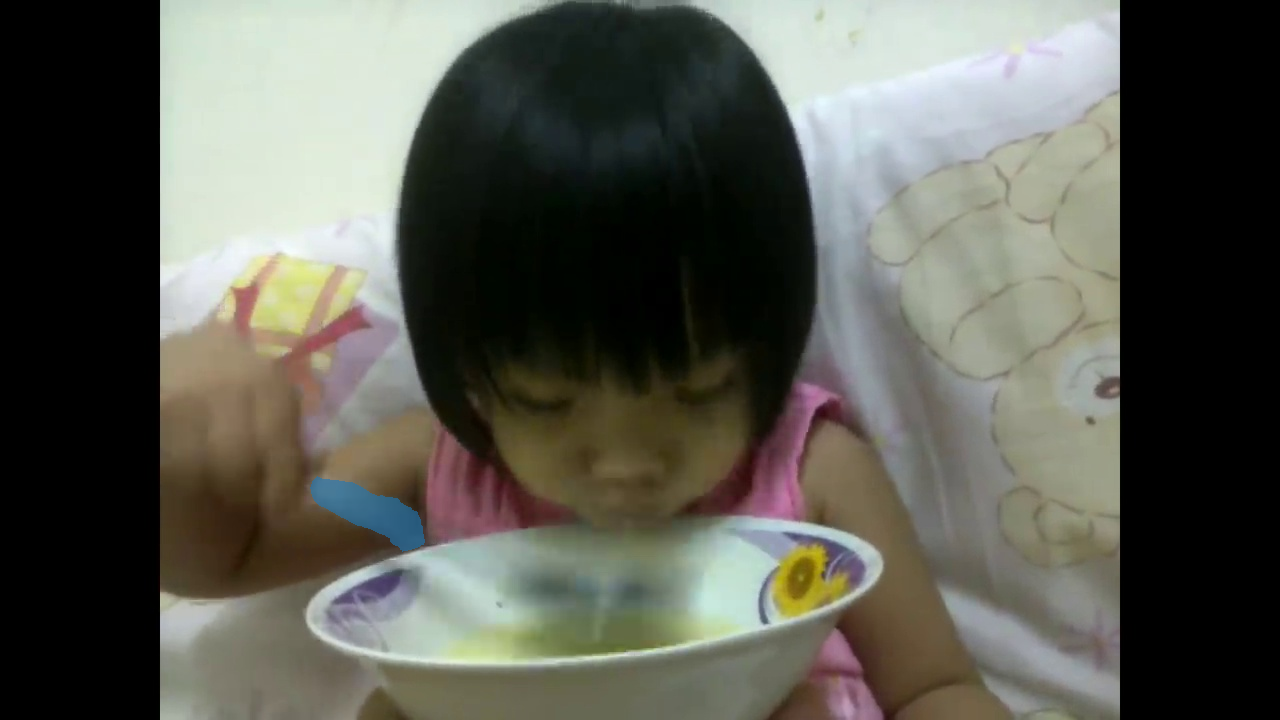} \\
    \end{tabular}
    \caption{Inaccurate segmentations, marked in blue, from Cutie and SAM2 due to motion blur.}
    \label{fig:fig_inaccurate_segmentation_spoon}
\end{figure}

The most common reason for inaccurate segmentations was motion blur. For spoons, a frequent issue was when a partially occluded spoon was lifted quickly, leading to sudden motion and subsequently segmentation errors. Examples of inaccurate segmentations are shown for both Cutie and SAM2 in \cref{fig:fig_inaccurate_segmentation_spoon}.

Two separate sequences are taken for Cutie and SAM2 here to show multiple examples of motion blur. It is relevant to note that SAM2 accurately segments the failure sequence shown for Cutie, but Cutie fails to segment the failure sequence used for SAM2.

Other common sources of segmentation error included misidentifying food or the bowl to be part of the spoon. Similar issues arose when segmenting the hand, such as segmenting the watch around the person's wrist or the entire arm as the hand like in \cref{fig:fig_occlusion_hand}.

\subsection{6D Pose Estimation}

With the raw RGB, depth and segmented frames obtained from previous steps, we use BundleSDF to obtain 6D pose estimates for each frame in the video sequence. 

\begin{figure}[h]
    \centering Cutie
    \vskip 0.2em
    \begin{tabular}{@{}c@{}c}
        \includegraphics[width=.5\linewidth]{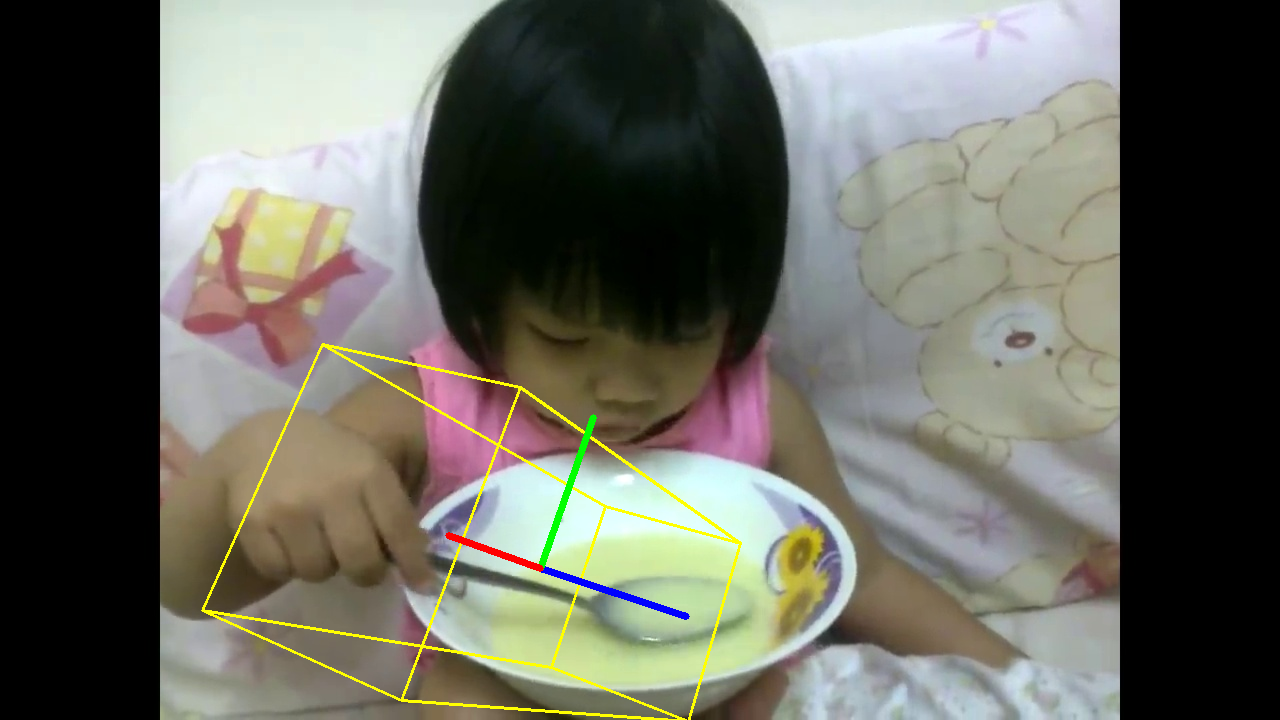} &
        \includegraphics[width=.5\linewidth]{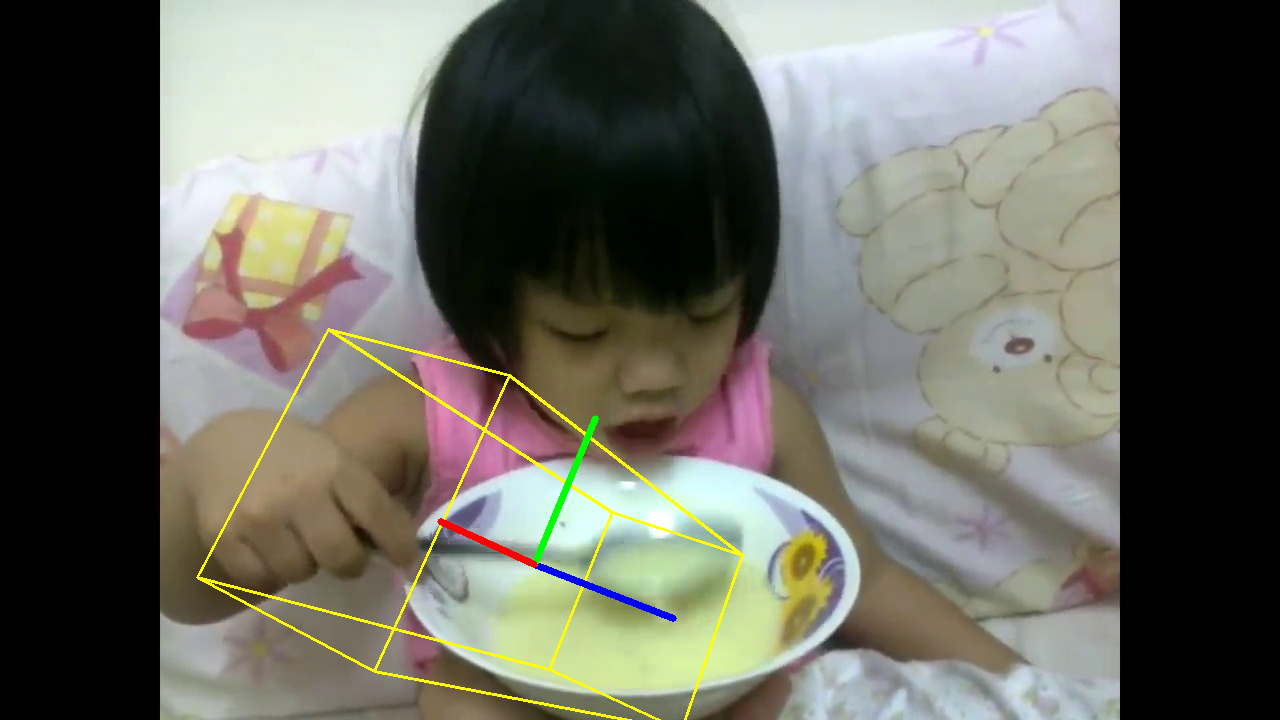} \\
        \includegraphics[width=.5\linewidth]{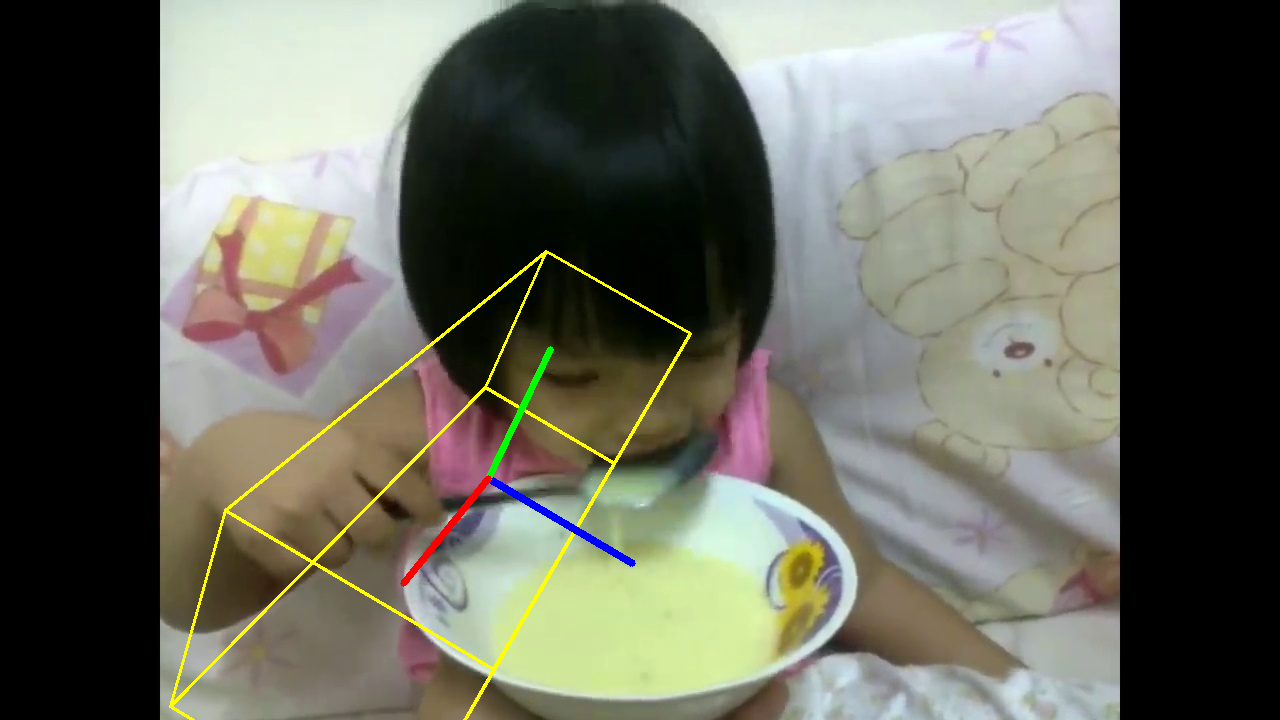} &
        \includegraphics[width=.5\linewidth]{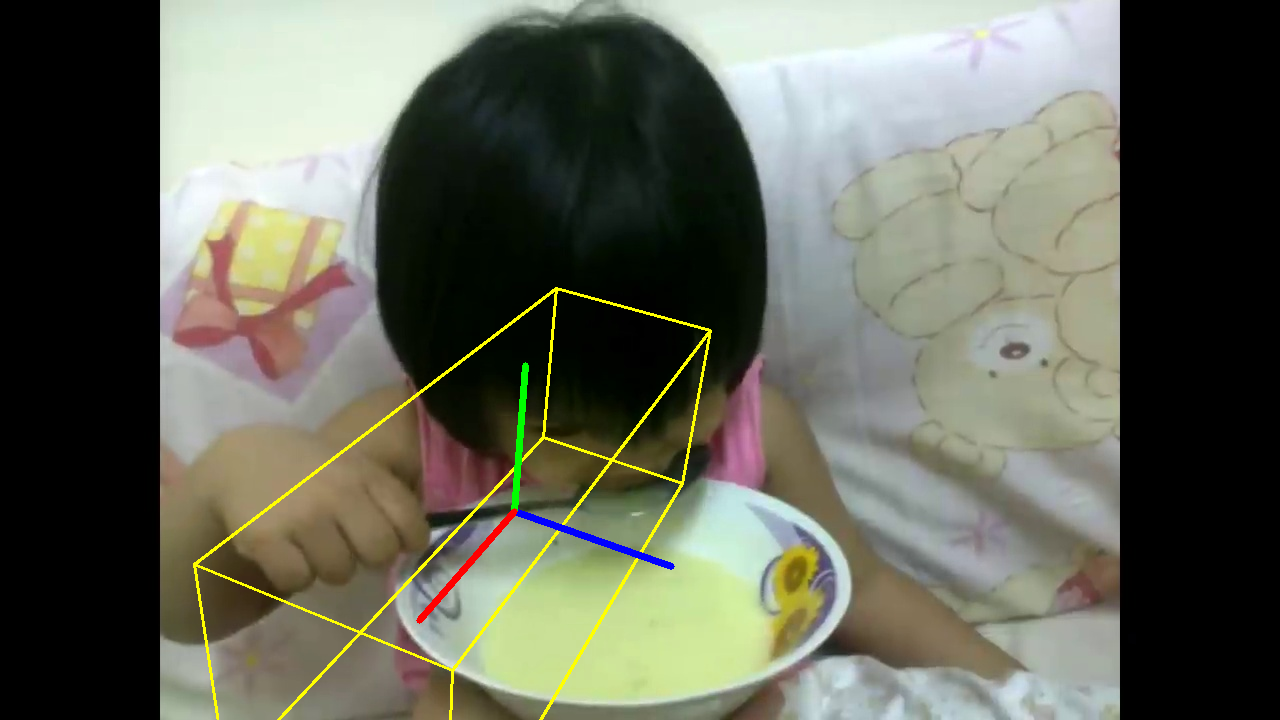} \\
    \end{tabular}
    \vskip 0.5em
    \centering SAM2
    \vskip 0.2em
    \begin{tabular}{@{}c@{}c}
        \includegraphics[width=.5\linewidth]{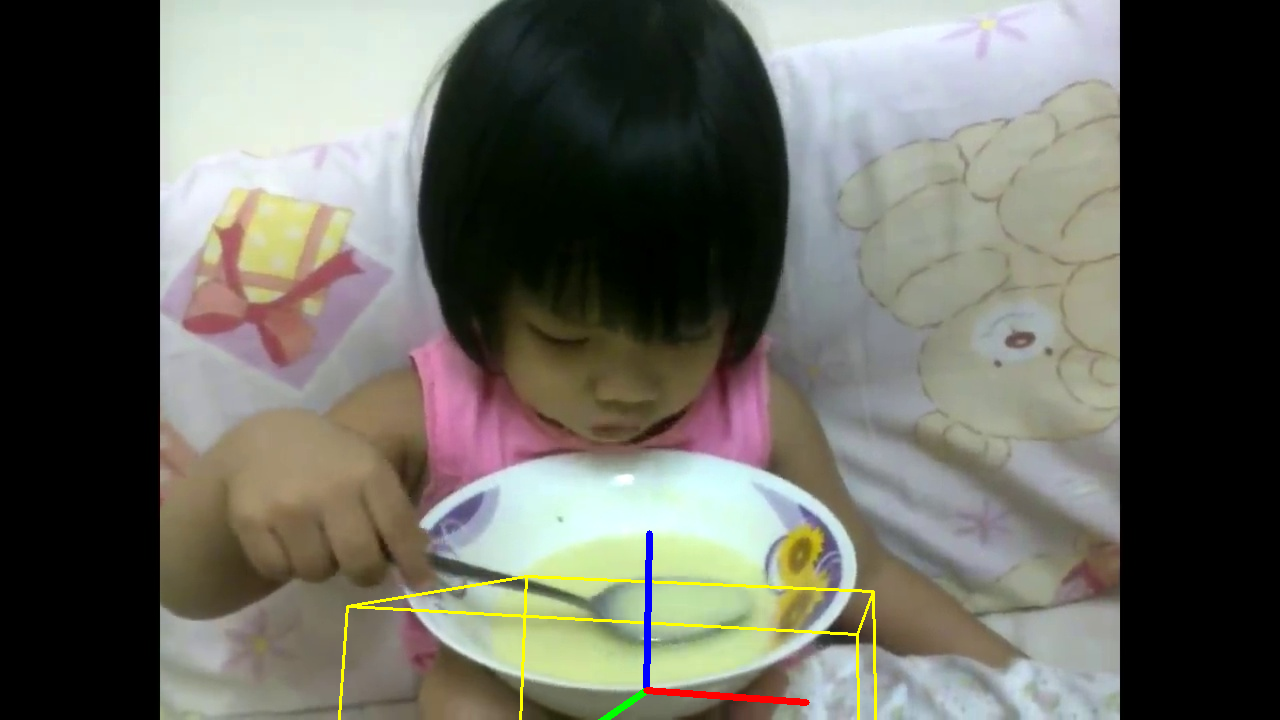} &
        \includegraphics[width=.5\linewidth]{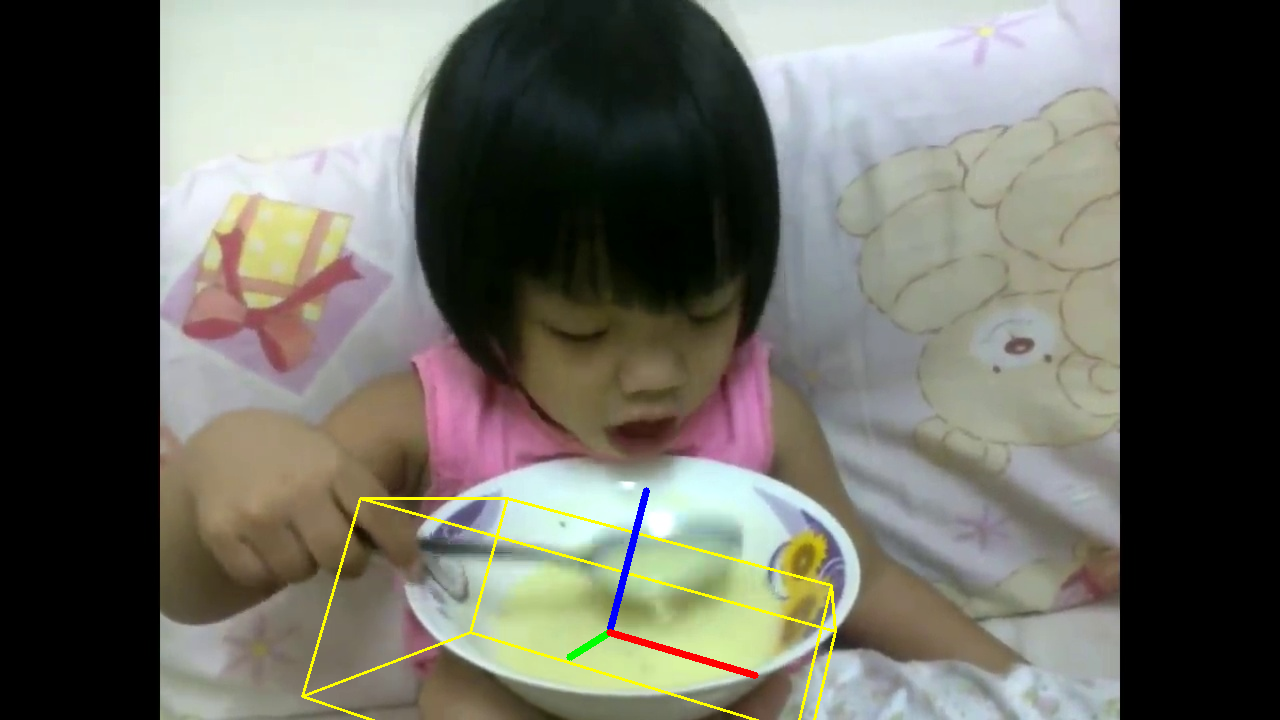} \\
        \includegraphics[width=.5\linewidth]{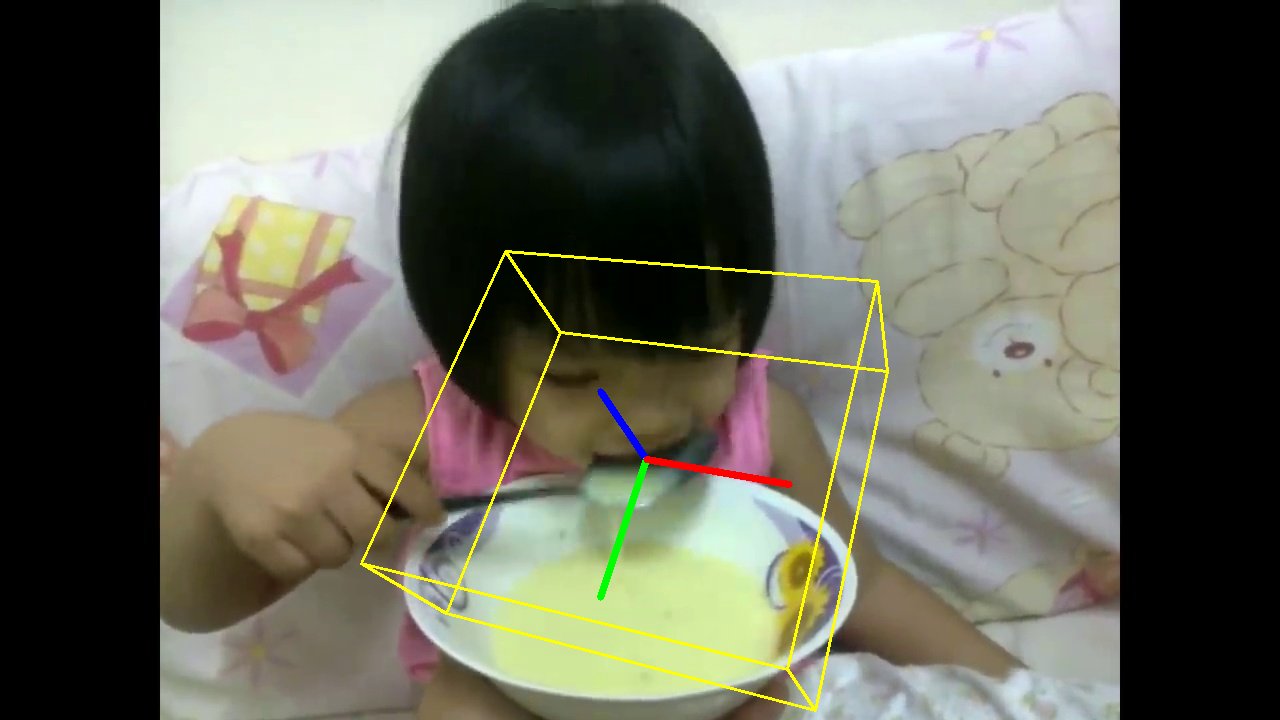} &
        \includegraphics[width=.5\linewidth]{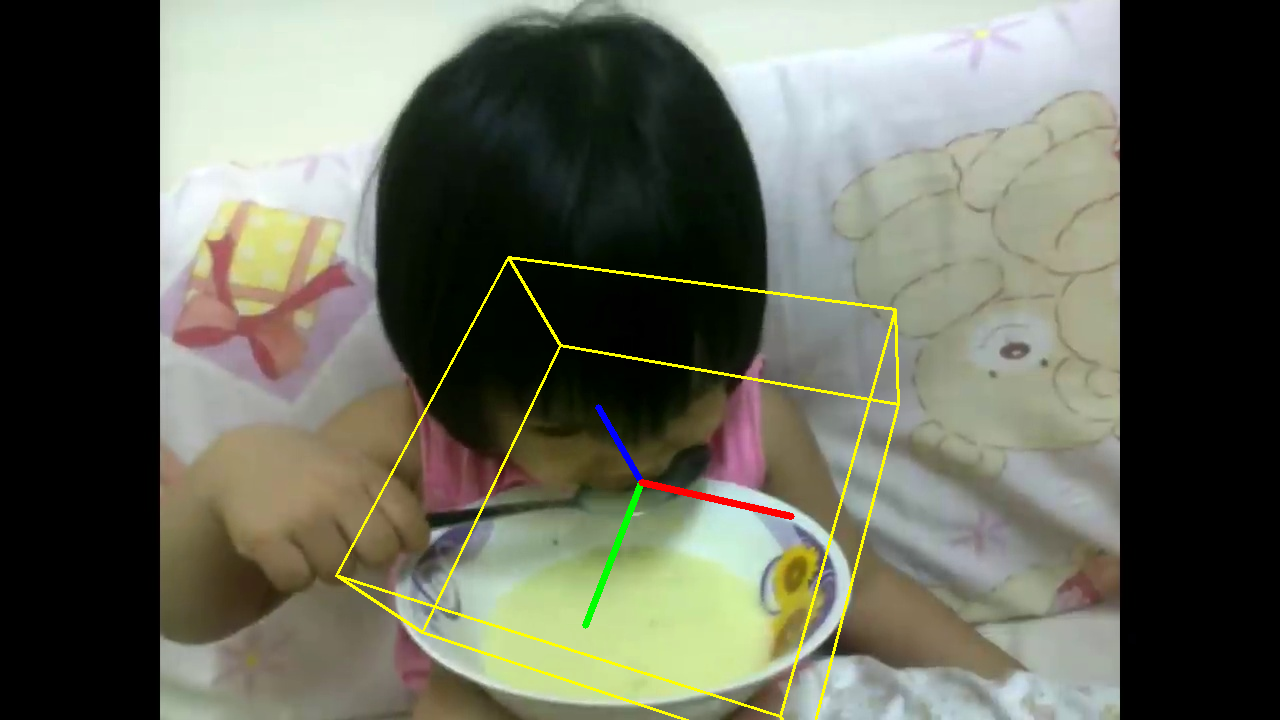} \\
    \end{tabular}
    \caption{6D Pose estimates for the spoon during an eating motion produced from segmented masks by both Cutie and SAM2.}
    \label{fig:fig_6d_pose_spoon}
\end{figure}

\begin{figure}[th]
    \centering Cutie
    \vskip 0.2em
    \begin{tabular}{@{}c@{}c}
        \includegraphics[width=.5\linewidth]{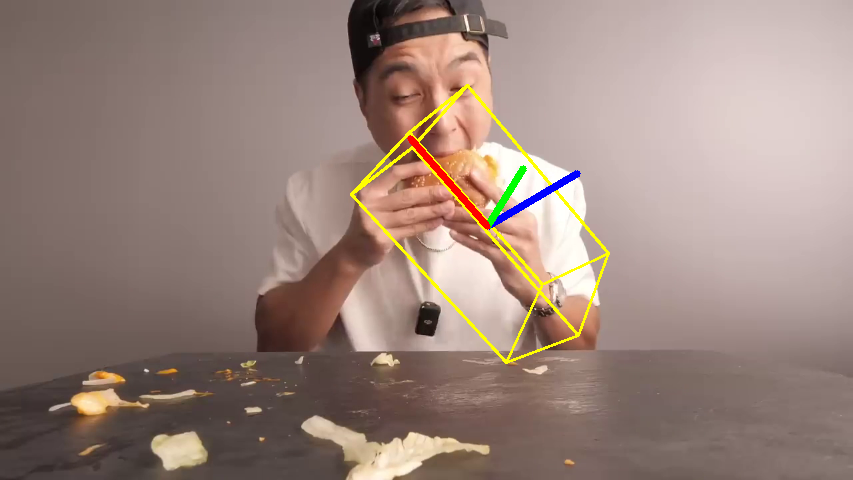} &
        \includegraphics[width=.5\linewidth]{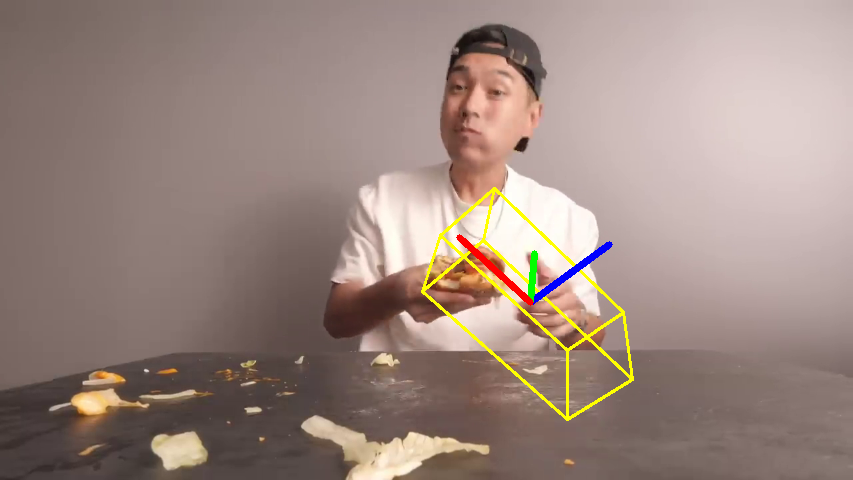} 
    \end{tabular}
    \vskip 0.5em
    \centering SAM2
    \vskip 0.2em
    \begin{tabular}{@{}c@{}c}
        \includegraphics[width=.5\linewidth]{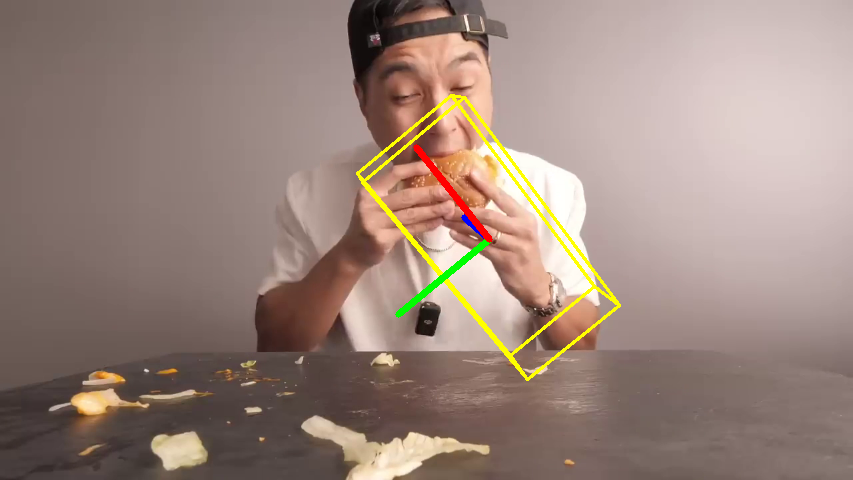} &
        \includegraphics[width=.5\linewidth]{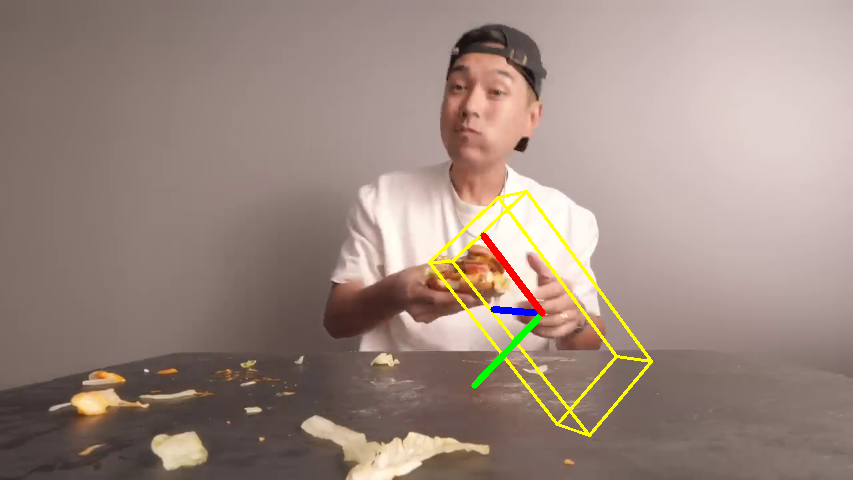}
    \end{tabular}
    \caption{6D Pose estimates for the hand on the right during an eating motion produced from segmented masks by both Cutie and SAM2.}
    \label{fig:fig_6d_pose_hand}
\end{figure}

BundleSDF is visually accurate in tracking the position of the objects throughout depth changes from hand movements or eating, as shown in \cref{fig:fig_6d_pose_spoon} and \cref{fig:fig_6d_pose_hand}.
It can be seen that the pitch, yaw, and roll estimates also remain consistent throughout the motion, even as the spoon enters the mouth and is partially occluded. 

\begin{figure}[h]
    \centering Cutie
    \vskip 0.2em
    \begin{tabular}{@{}c@{}c}
        \includegraphics[width=.5\linewidth]{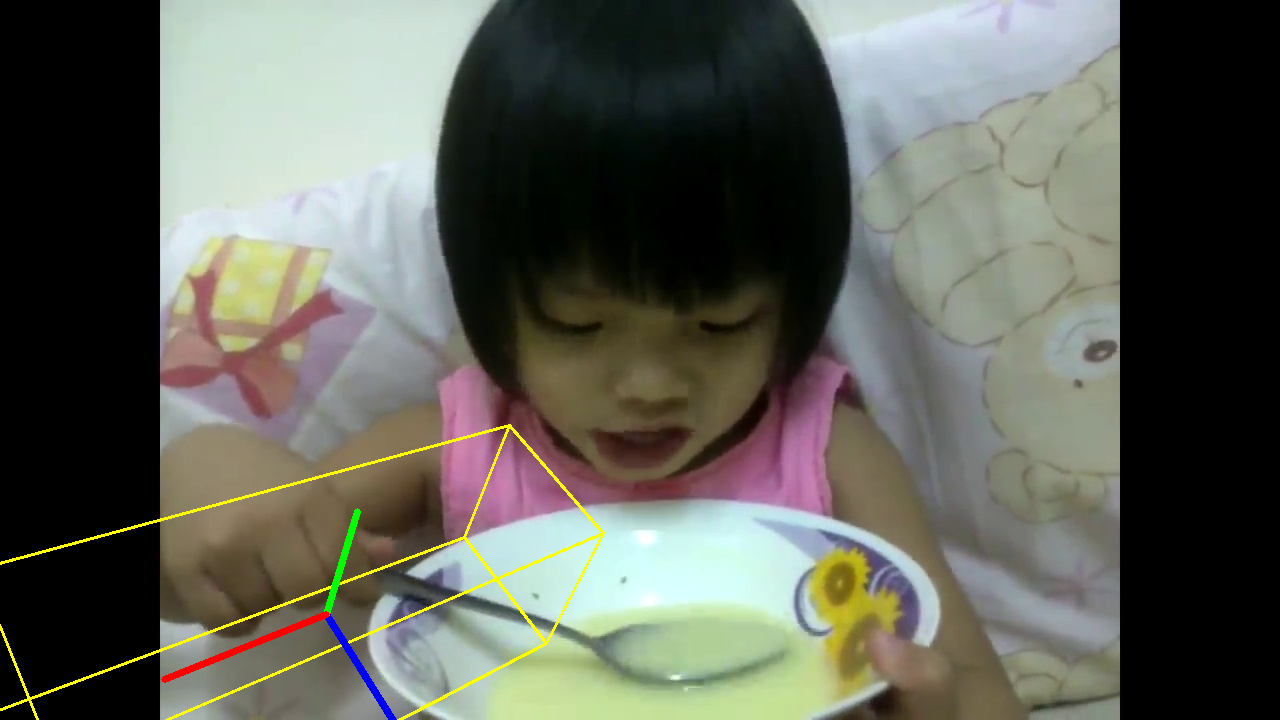} &
        \includegraphics[width=.5\linewidth]{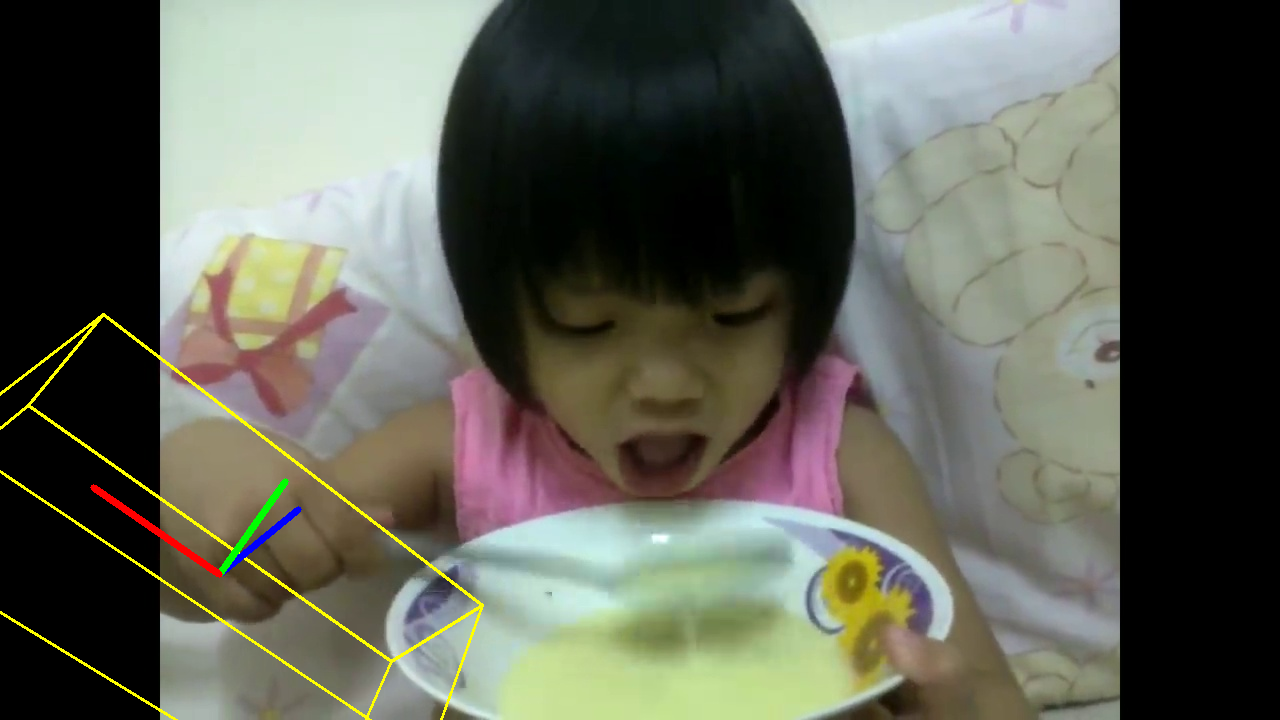} \\
    \end{tabular}
    \vskip 0.5em
    \centering SAM2
    \vskip 0.2em    
    \begin{tabular}{@{}c@{}c}
        \includegraphics[width=.5\linewidth]{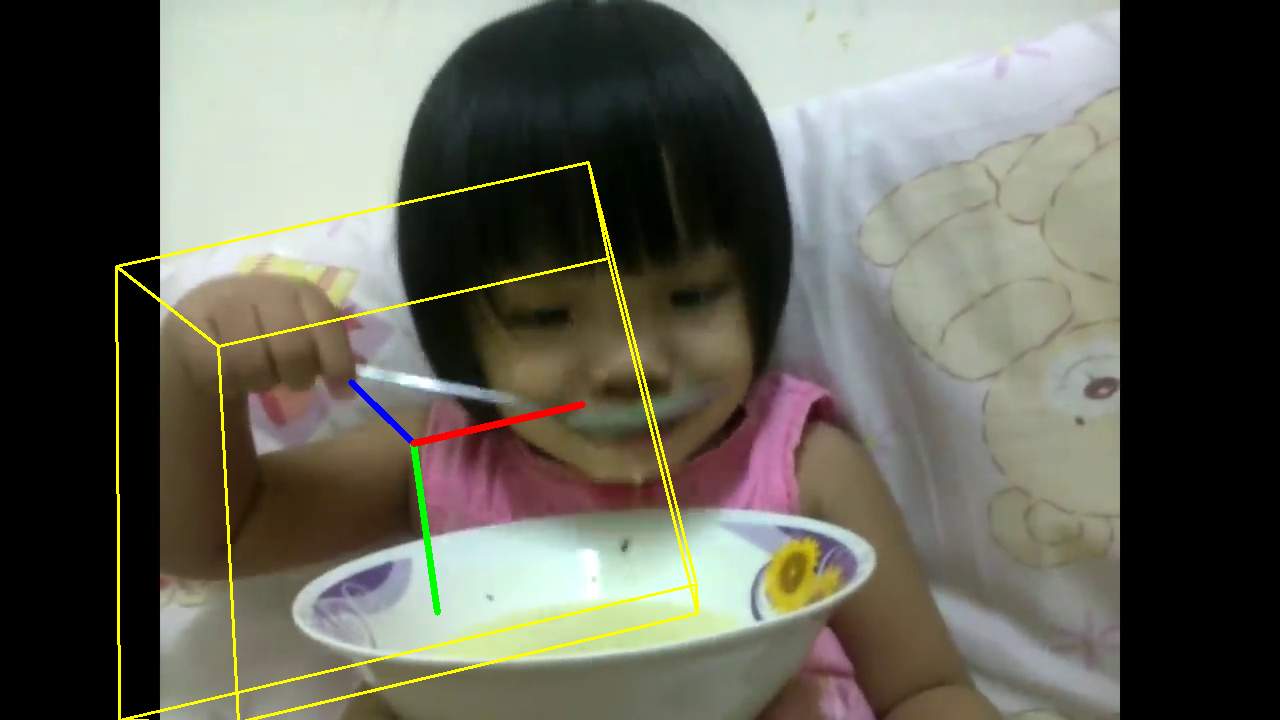} &
        \includegraphics[width=.5\linewidth]{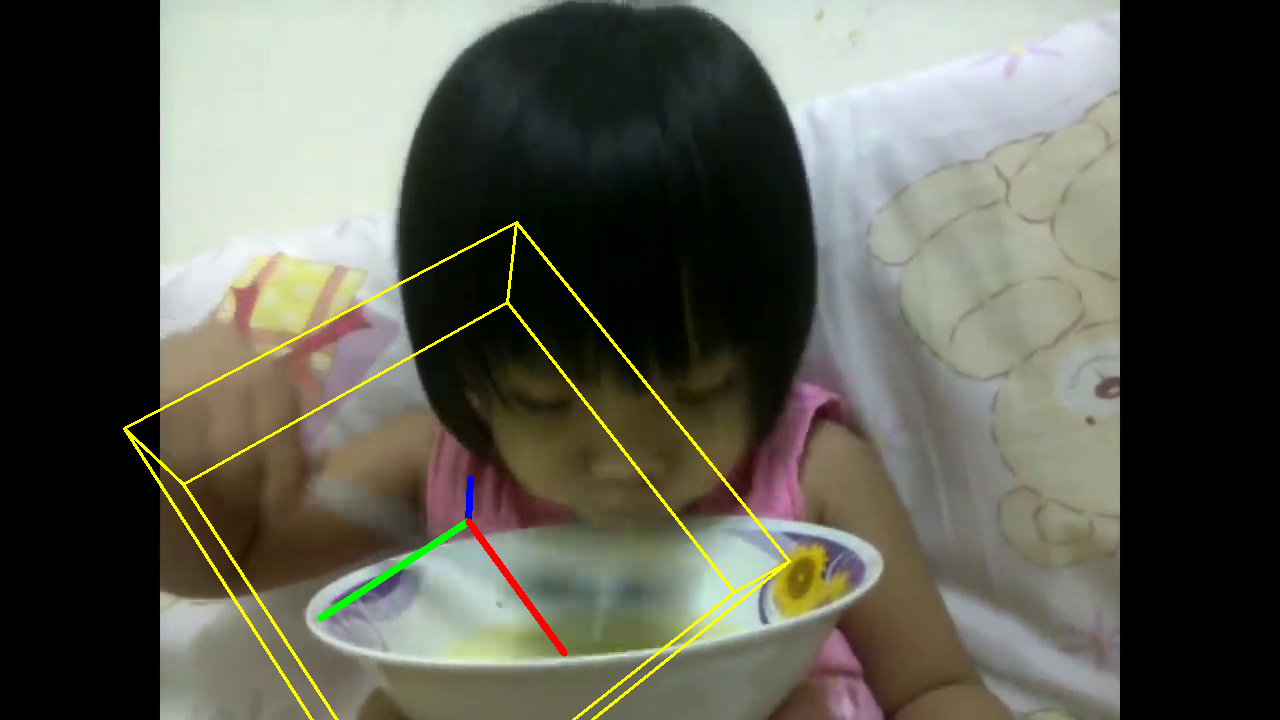} \\
    \end{tabular}
    \caption{BundleSDF is able to track objects despite poor segmentation.}
    \label{fig:fig_inaccurate_segmentation_good_pose_spoon}
\end{figure}

It is especially interesting to investigate how BundleSDF performs in instances of inaccurate segmentation. In the examples of segmentation failures from the previous section, BundleSDF is still able to track the objects successfully for SAM2 with a consistent orientation estimate, but is unable to correctly assess the position of the spoon with segmentations from Cutie, as shown in \cref{fig:fig_inaccurate_segmentation_good_pose_spoon}.

For spoons, VOS with Cutie leads to more inaccurate pose estimations than VOS with SAM2. 

BundleSDF is also able to provide consistent and accurate rotational estimates for the hand in the process of eating, as shown in \cref{fig:fig_6d_pose_hand}. 

Overall, for both spoons and hands, BundleSDF using segmentation masks from either Cutie or SAM2 is able to perform 6D pose estimation accurately for general food consumption. However, it may fall short in more complex scenarios.

\subsection{Robustness to Occlusion}

BundleSDF can have difficulty to track objects after they are partially or fully occluded. 
\begin{figure}[h]
    \centering Cutie
    \vskip 0.2em
    \begin{tabular}{@{}c@{}c@{}c}
         \includegraphics[width=.333\linewidth]{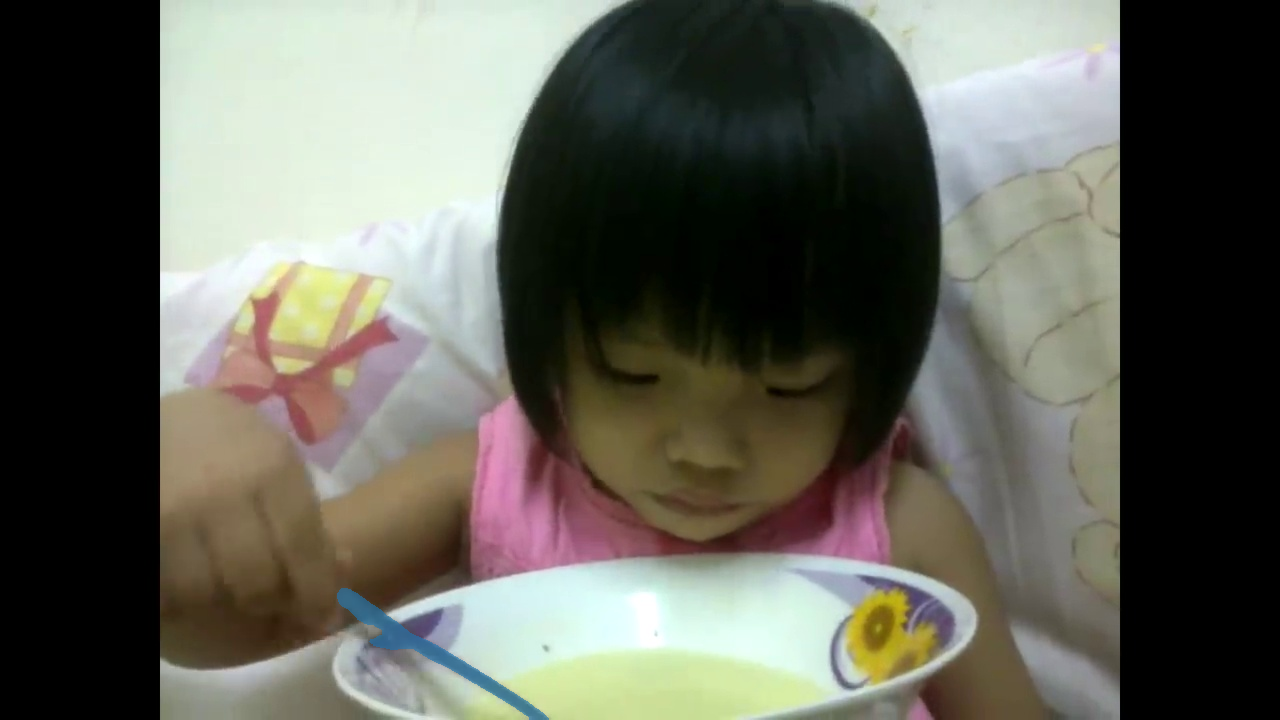} &
        \includegraphics[width=.333\linewidth]{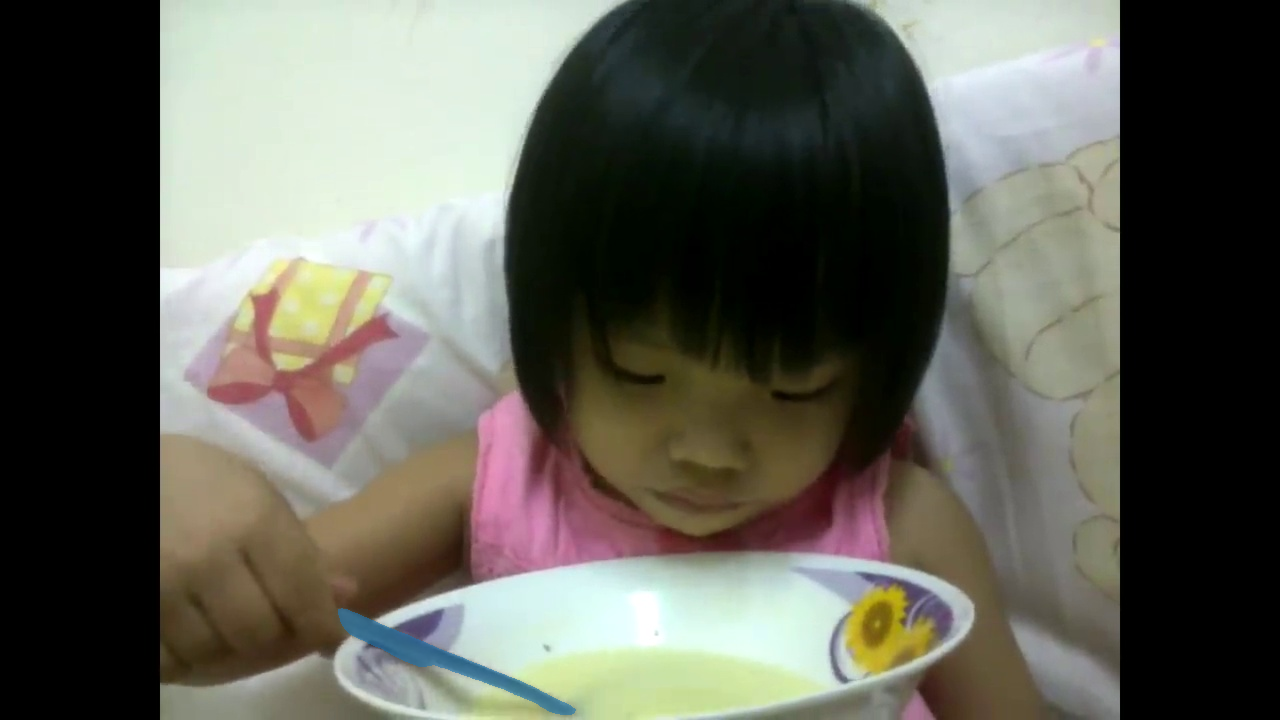} &
        \includegraphics[width=.333\linewidth]{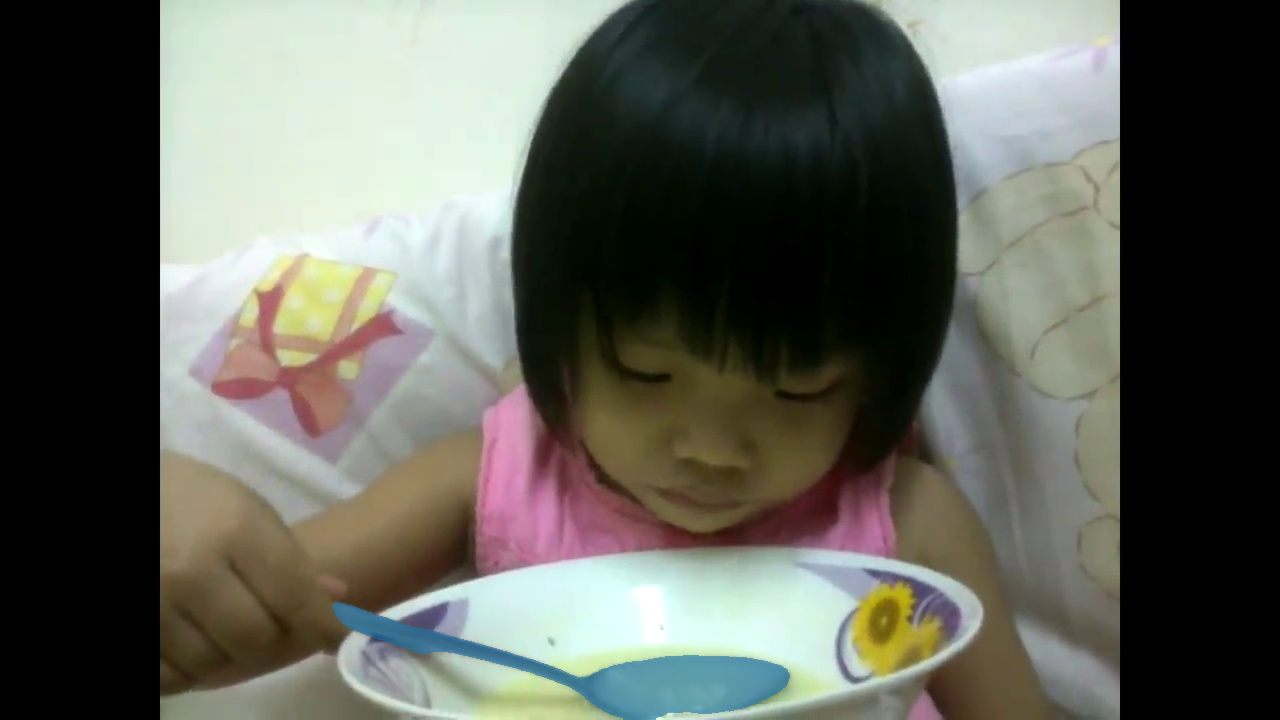} \\
        \includegraphics[width=.333\linewidth]{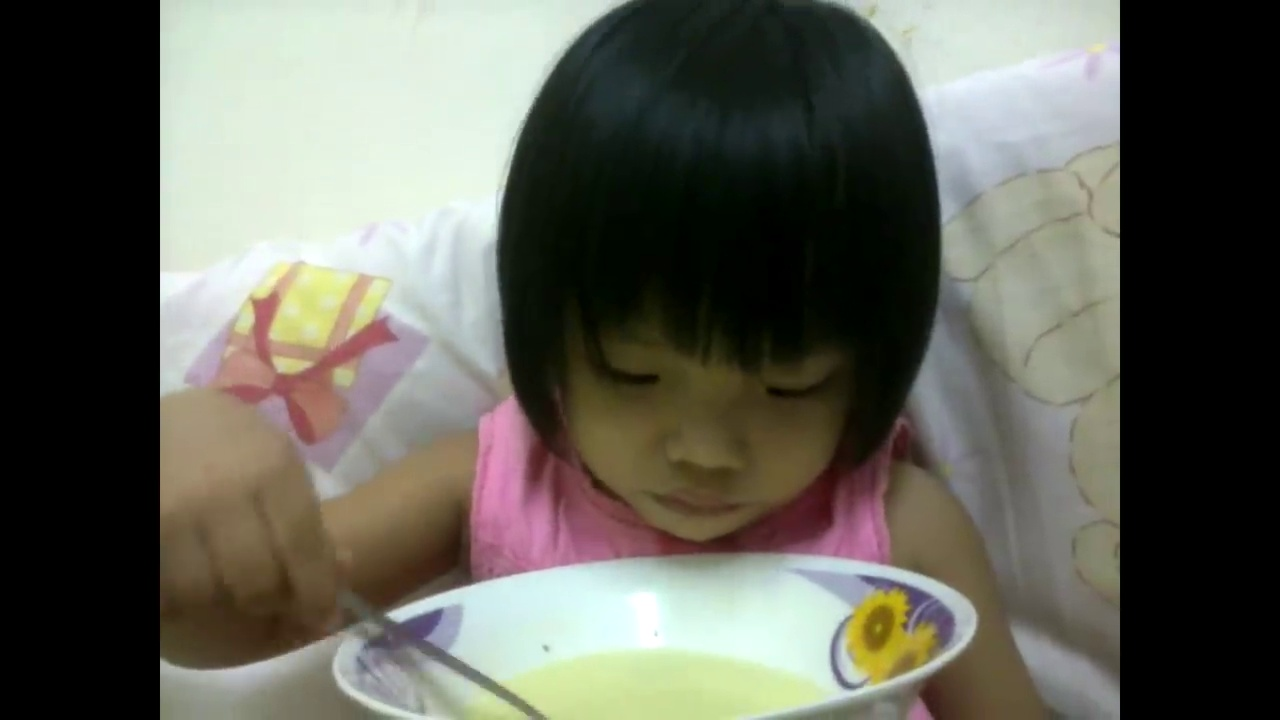} &
        \includegraphics[width=.333\linewidth]{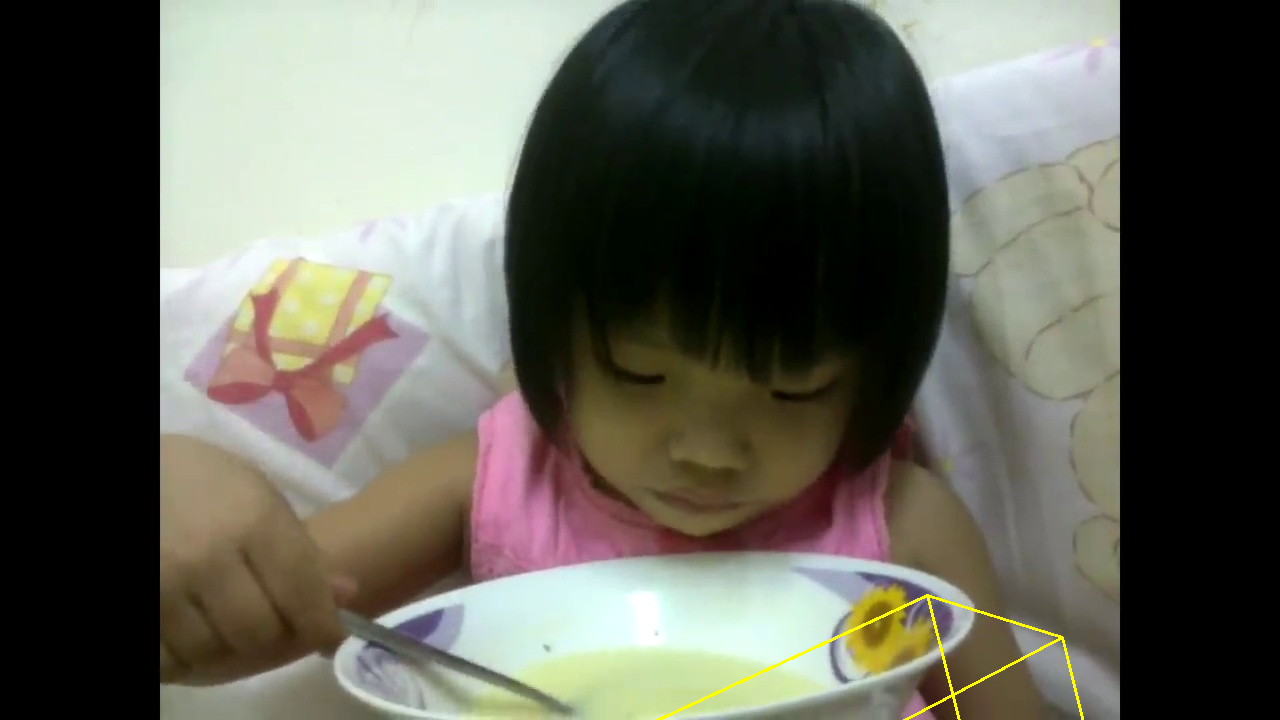} &
        \includegraphics[width=.333\linewidth]{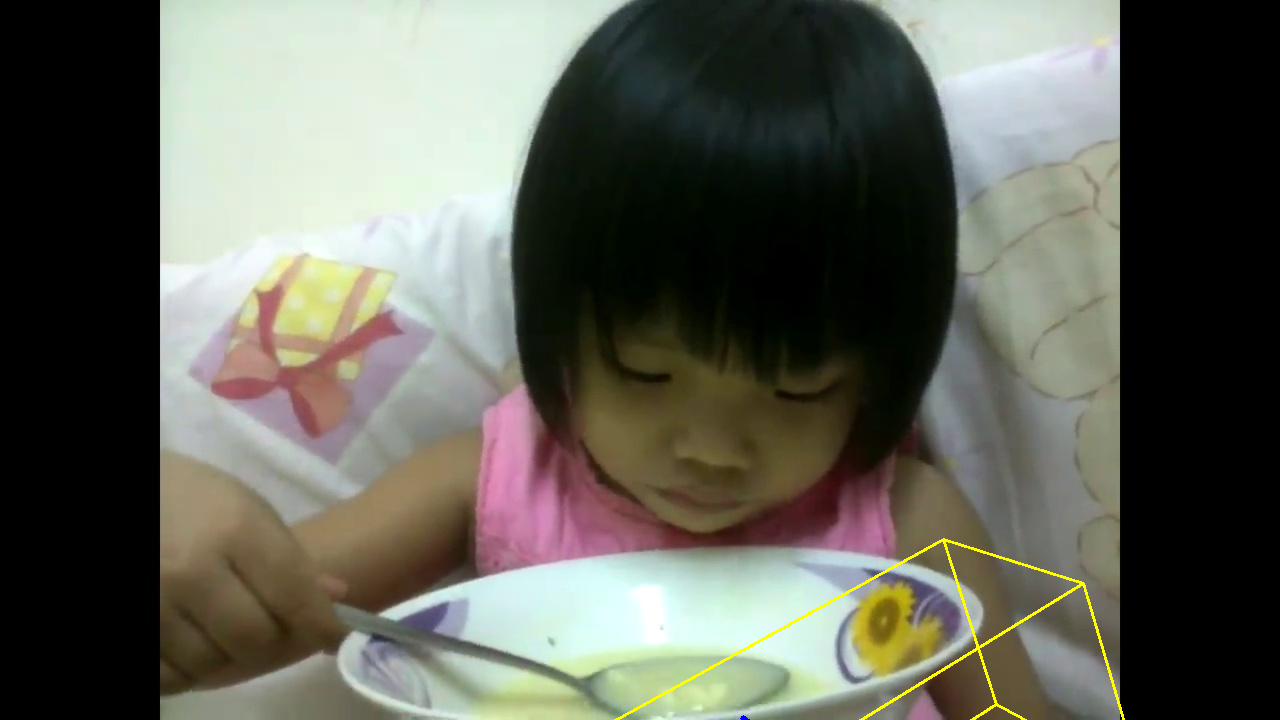} \\
    \end{tabular}
    \vskip 0.5em
    \centering SAM2
    \vskip 0.2em
    \begin{tabular}{@{}c@{}c@{}c}
        \includegraphics[width=.333\linewidth]{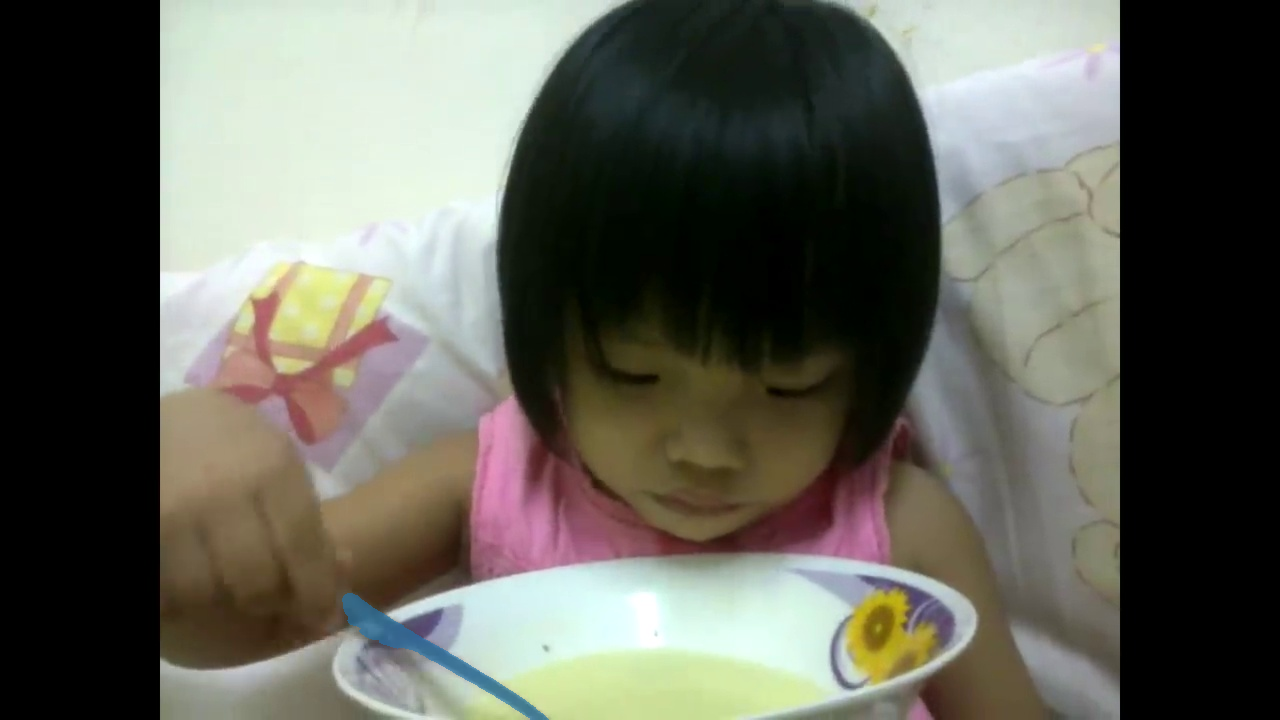} &
        \includegraphics[width=.333\linewidth]{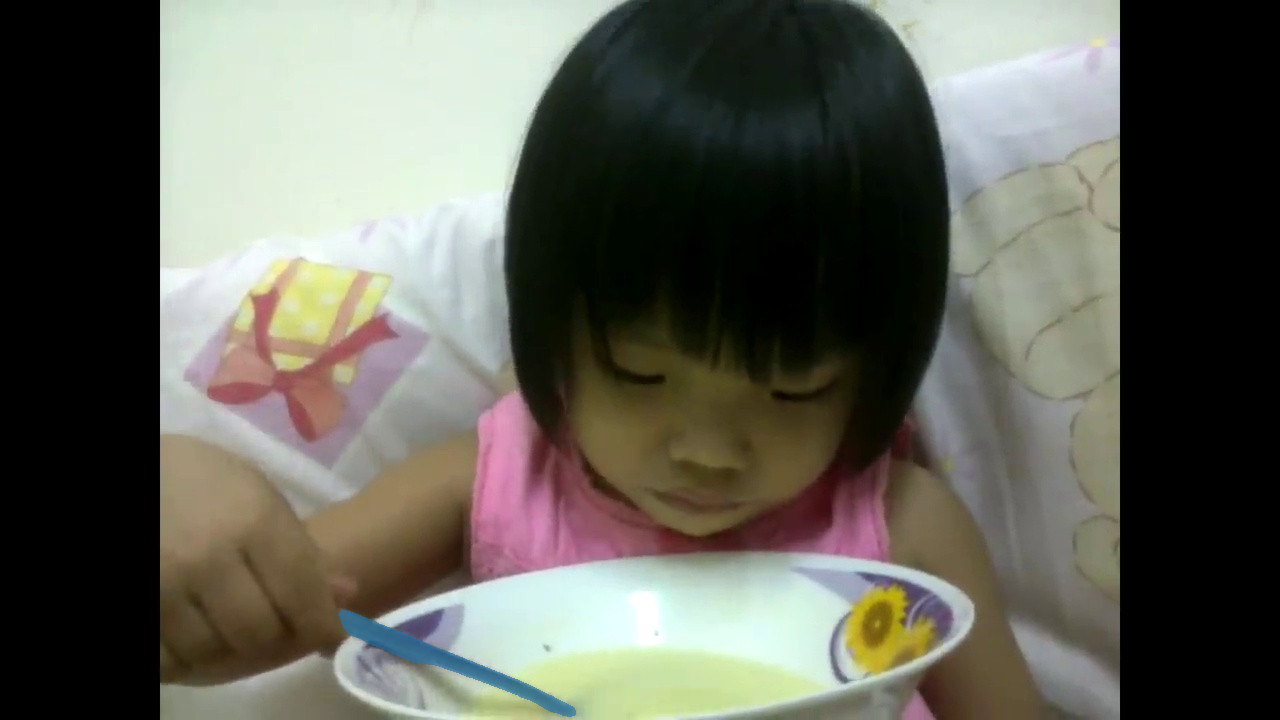} &
        \includegraphics[width=.333\linewidth]{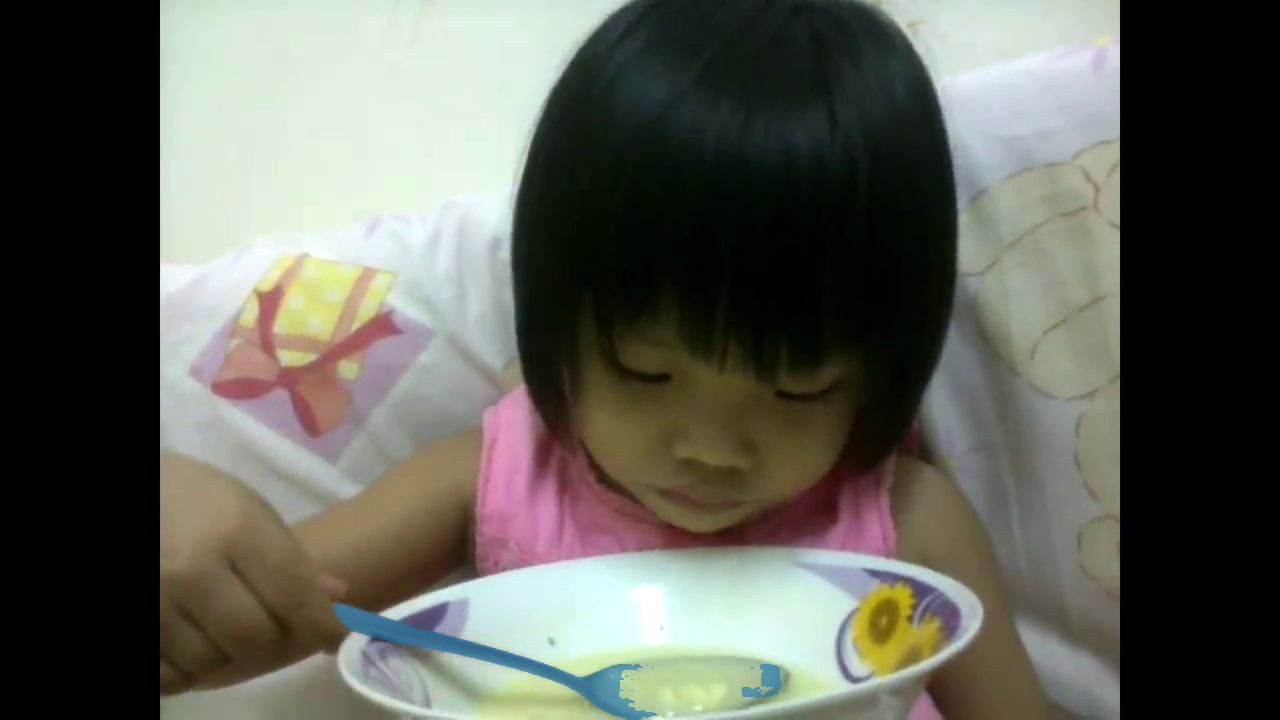} \\
        \includegraphics[width=.333\linewidth]{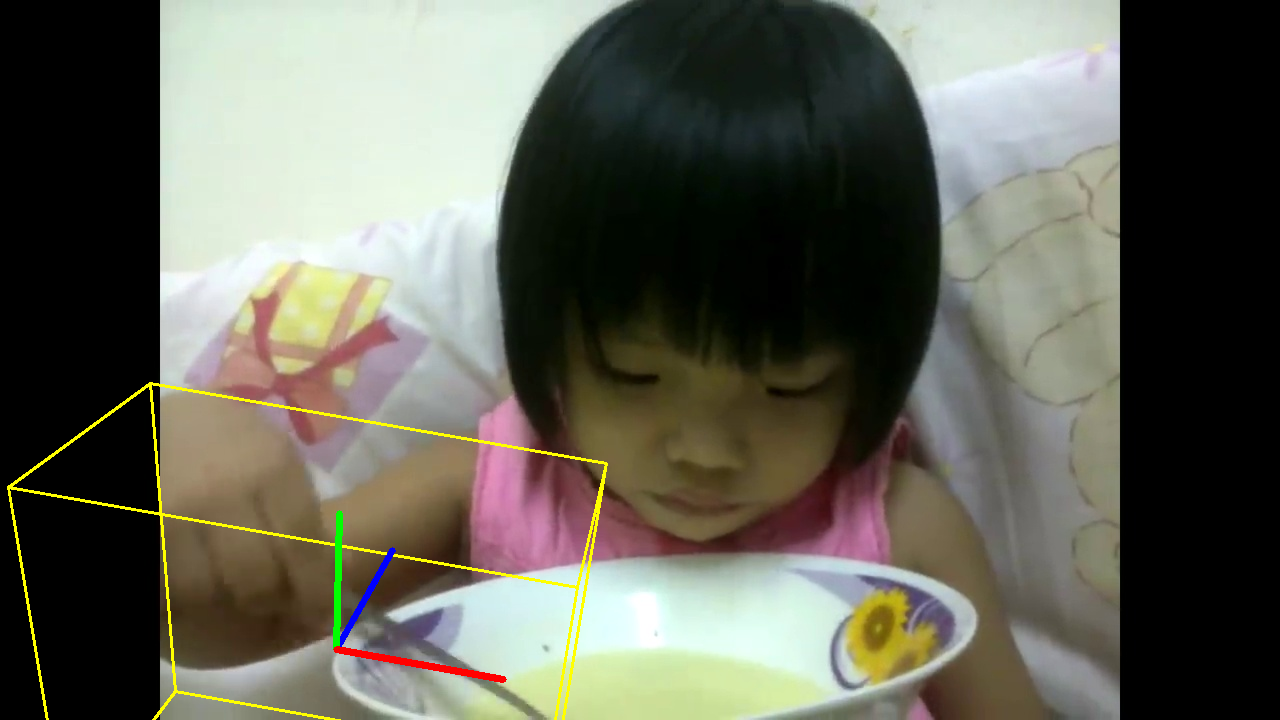} &
        \includegraphics[width=.333\linewidth]{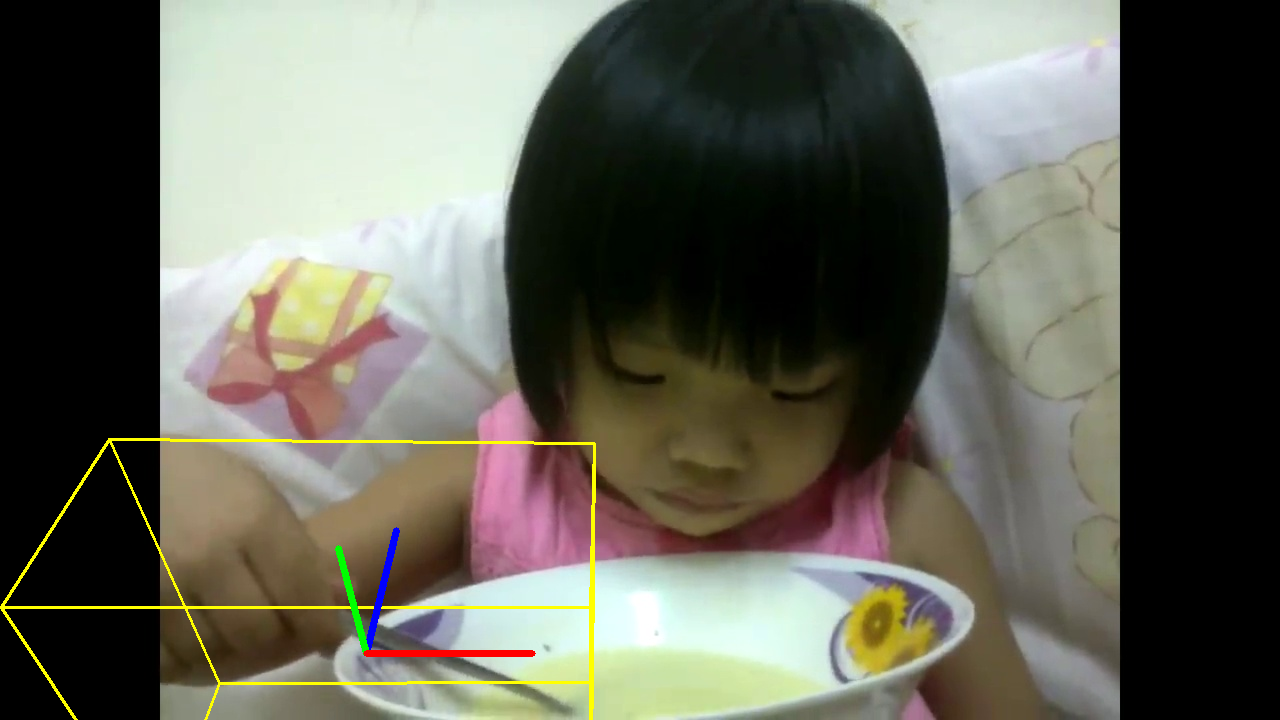} &
        \includegraphics[width=.333\linewidth]{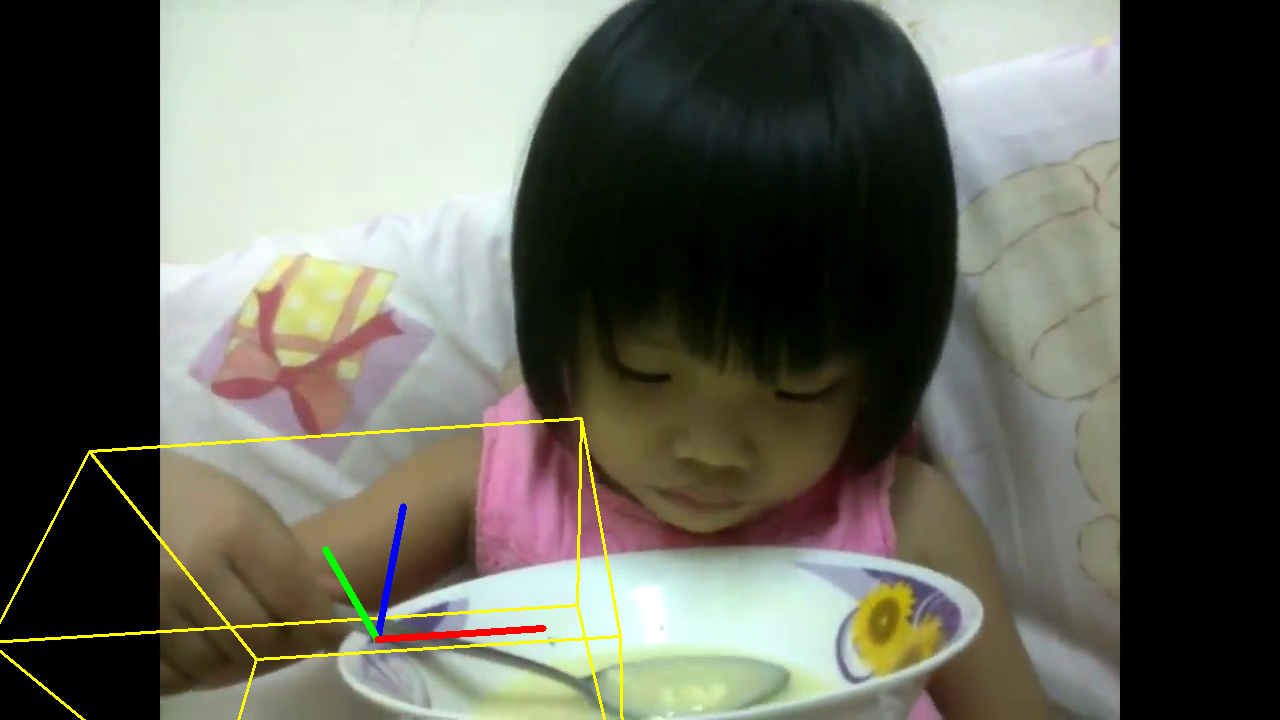} \\
    \end{tabular}
    \caption{First row for both VOS models shows object segmentation. Second row shows pose estimation.}
    \label{fig:fig_occlusion_spoon}
\end{figure}
For example, when the spoon scoop is occluded by the soup in \cref{fig:fig_occlusion_spoon}, both VOS models are able to accurately segment the handle of the spoon, as well as the scoop when the spoon surfaces. However, BundleSDF's pose estimation is incorrect for the entire duration of the sequence in Cutie's case, and does not update to reflect the spoon surfacing in SAM2's case.

\begin{figure}[th]
    \begin{tabular}{@{}c@{}c}
         \includegraphics[width=.5\linewidth]{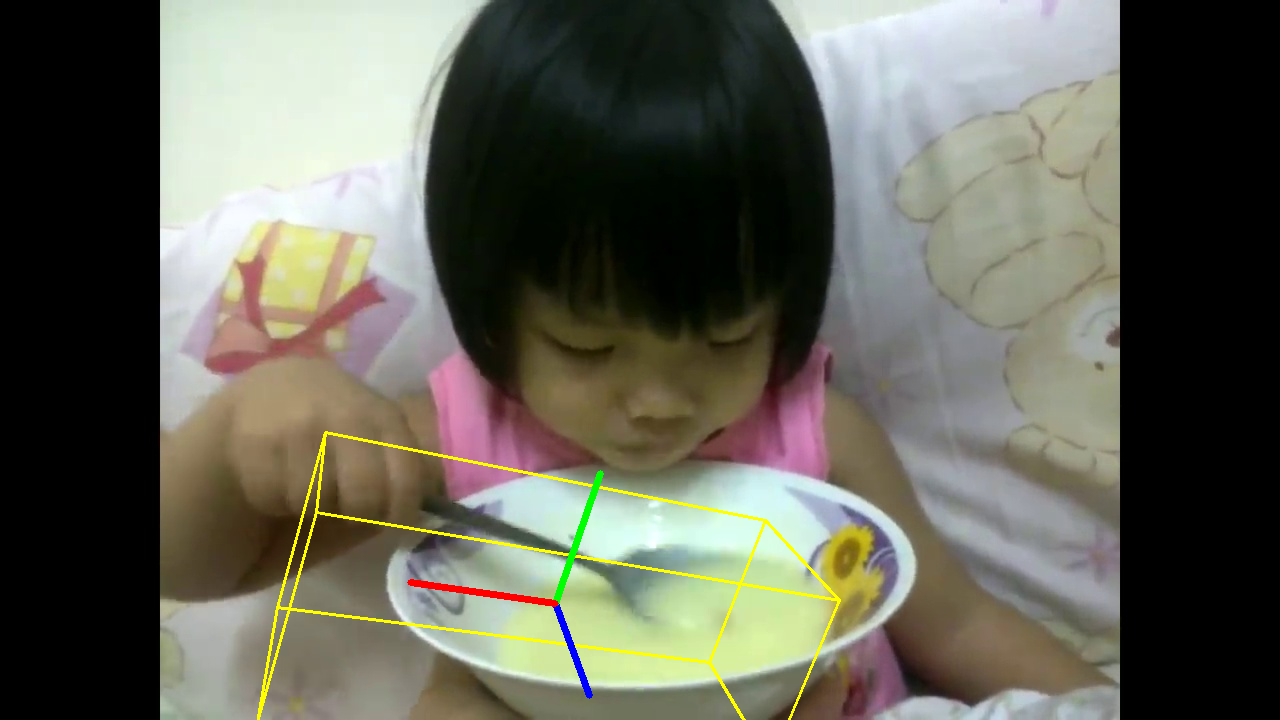} &
        \includegraphics[width=.5\linewidth]{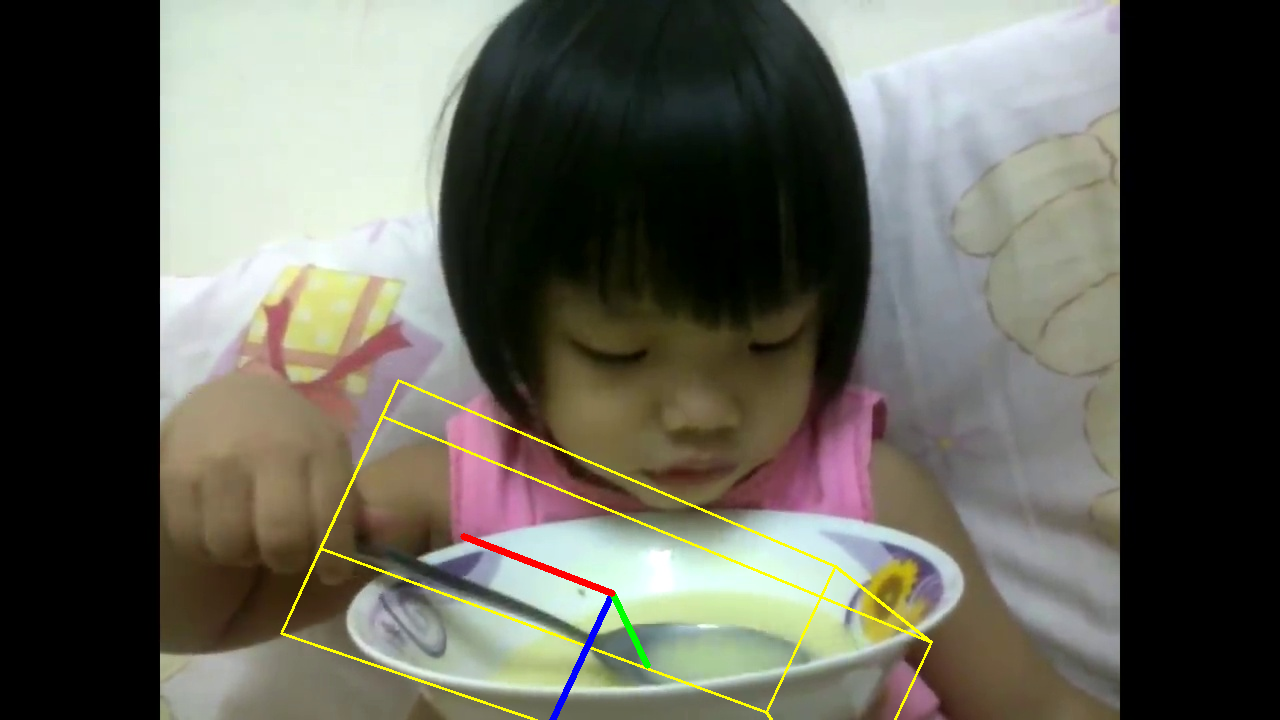} \\
    \end{tabular}
    \caption{BundleSDF orientation estimates flipped across frames, despite the spoon's orientation being similar.}
    \label{fig:fig_inconsistent_orientation_spoon}
\end{figure}

The object being occluded may cause BundleSDF to either estimate a visibly incorrect bounding box for the object, as above, or estimate a rotational orientation inconsistent with previous frames, as shown in \cref{fig:fig_inconsistent_orientation_spoon}.

\begin{figure}[h]
    \centering
    Cutie
    \vskip 0.2em
    \begin{tabular}{@{}c@{}c@{}c}
         \includegraphics[width=.333\linewidth]{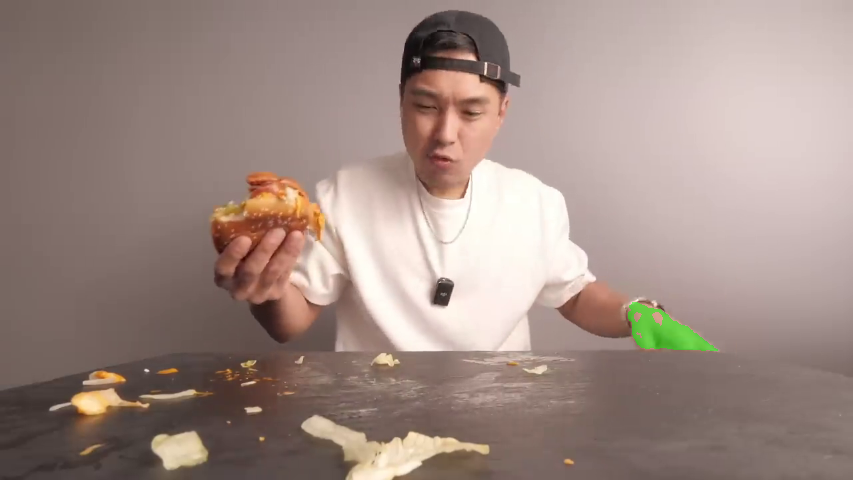} &
        \includegraphics[width=.333\linewidth]{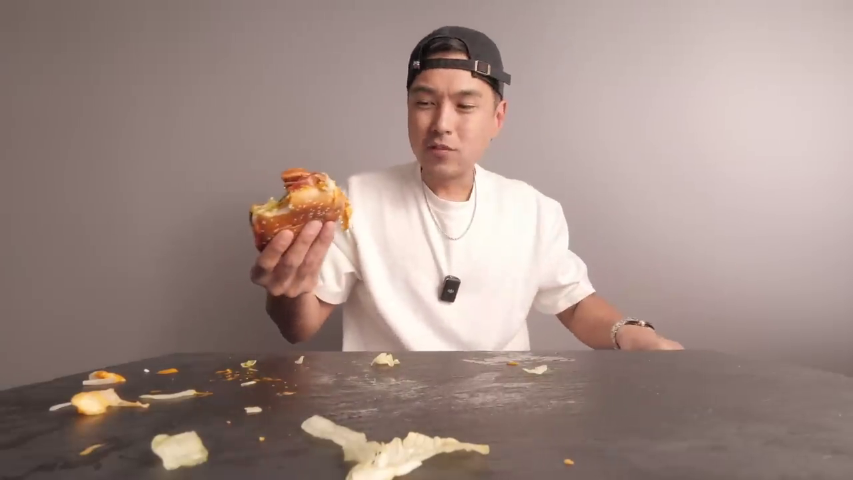} &
        \includegraphics[width=.333\linewidth]{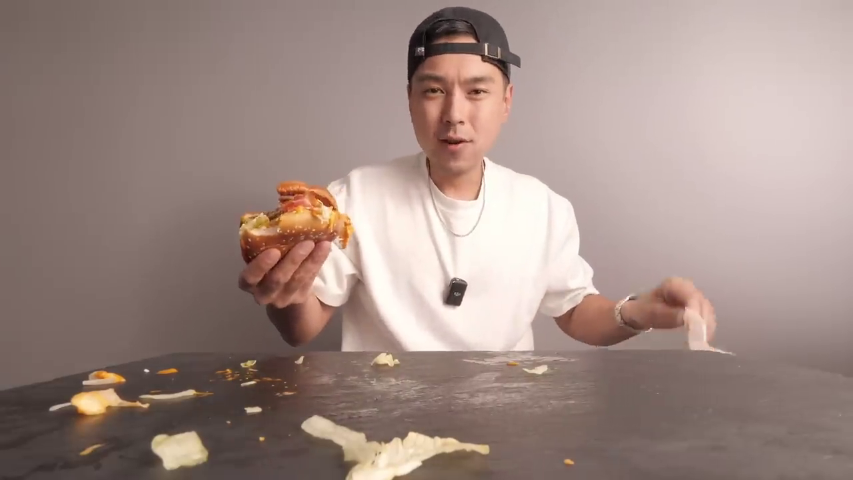} \\
        \includegraphics[width=.333\linewidth]{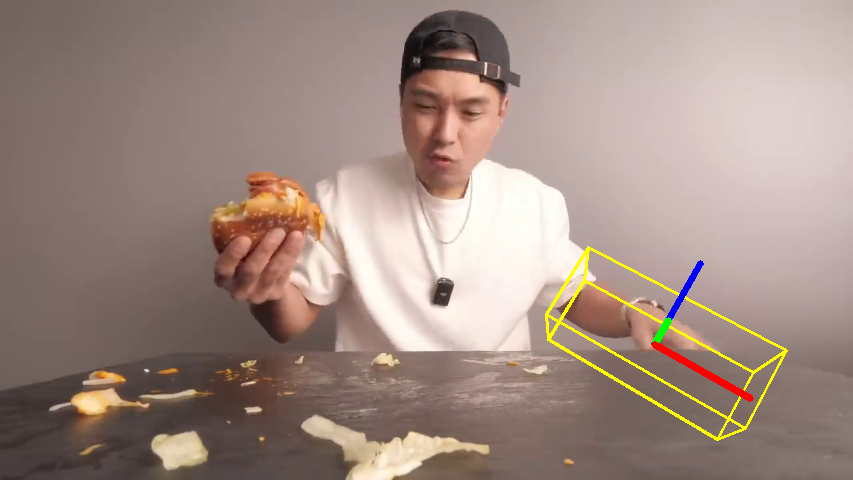} &
        \includegraphics[width=.333\linewidth]{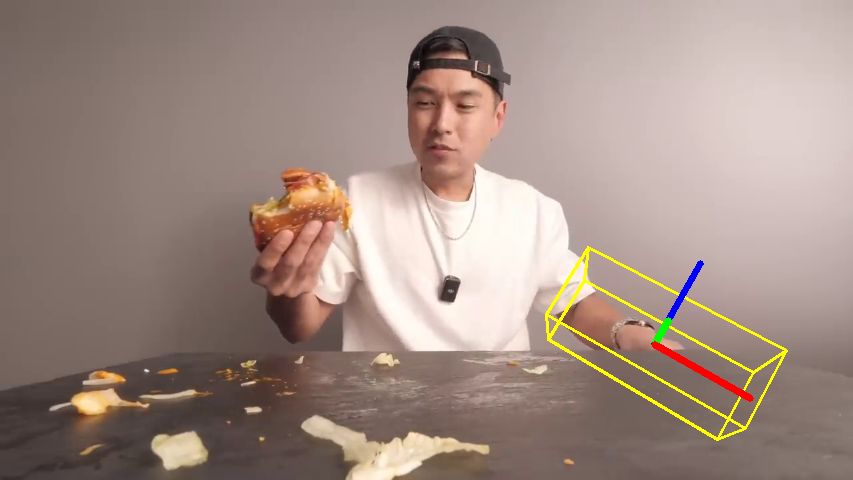} &
        \includegraphics[width=.333\linewidth]{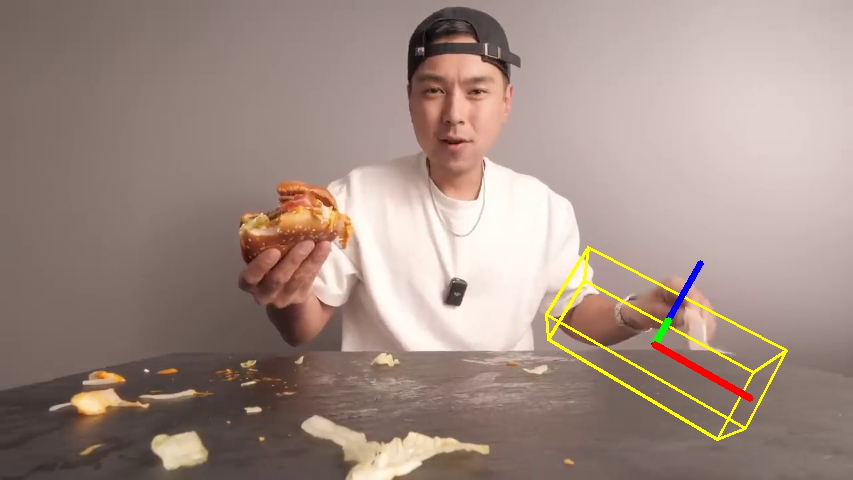} \\
    \end{tabular}
    \vskip 0.5em
    SAM2
    \vskip 0.2em
    \begin{tabular}{@{}c@{}c@{}c}
         \includegraphics[width=.333\linewidth]{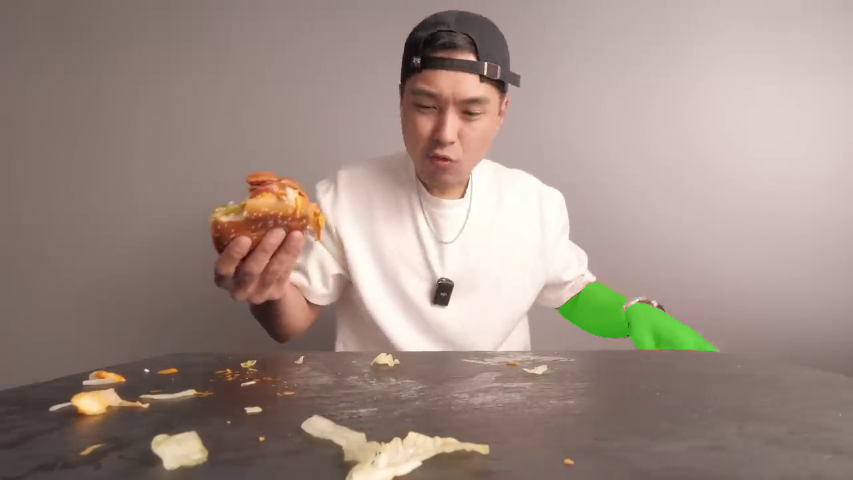} &
        \includegraphics[width=.333\linewidth]{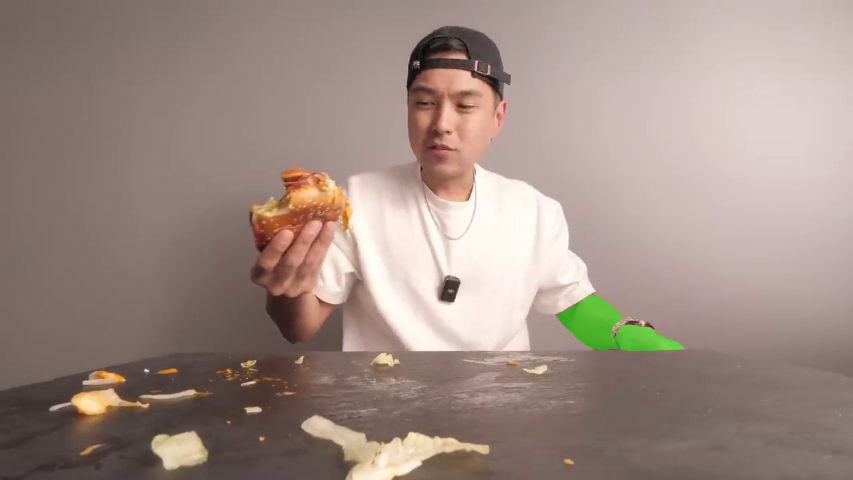} &
        \includegraphics[width=.333\linewidth]{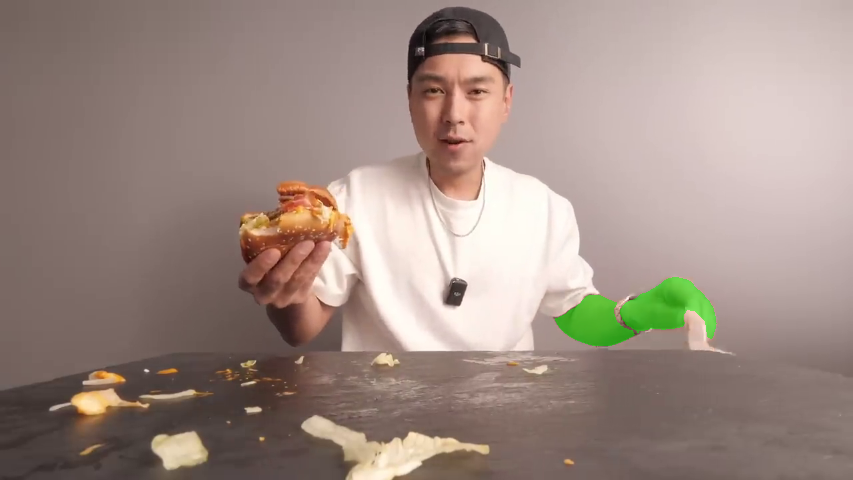} \\
        \includegraphics[width=.333\linewidth]{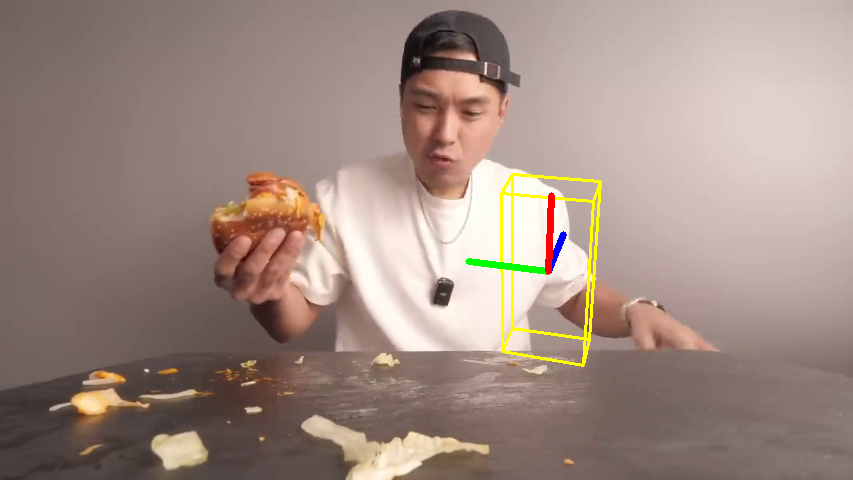} &
        \includegraphics[width=.333\linewidth]{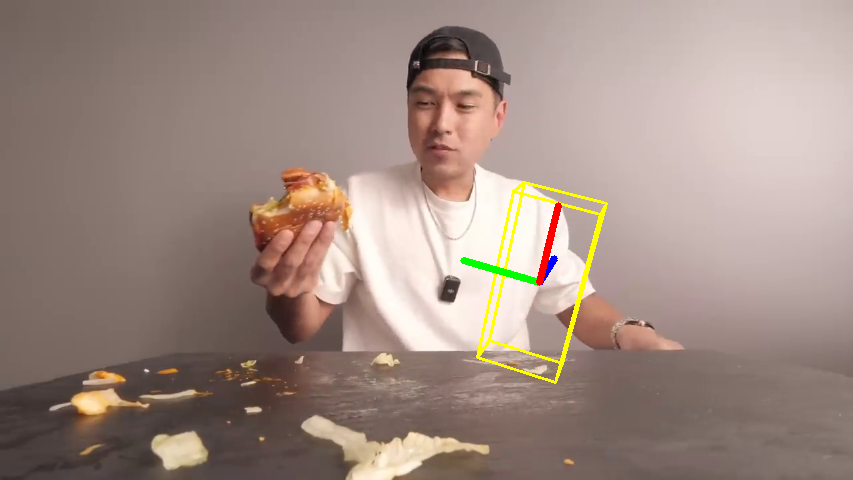} &
        \includegraphics[width=.333\linewidth]{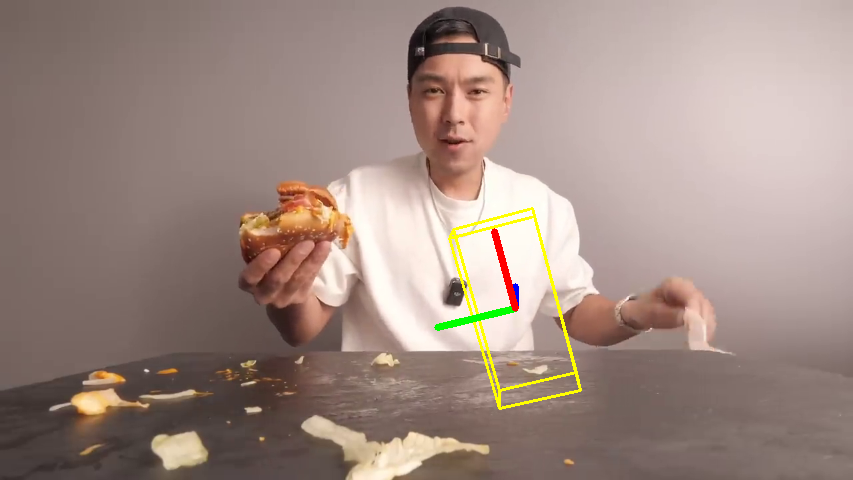} \\
    \end{tabular}
    \caption{Example for hand occlusion. For each VOS model, the first row shows object segmentation and the second row shows the pose estimation.}
    \label{fig:fig_occlusion_hand}
\end{figure}

When segmenting the hand, Cutie appears to perform worse when faced with occlusions as seen in \cref{fig:fig_occlusion_hand}. In this case, the hand goes under the table, causing Cutie to lose track of it. As a result, BundleSDF similarly loses track and is unable to adjust according to the change in pose of the hand. In the same frames, SAM2 is able to keep track of the hand even after going under the table, but starts to be inaccurate by segmenting the entire arm rather than just the right hand. The effects of this inaccuracy is significant, causing BundleSDF to generate a visibly incorrect bounding box.

\section{Conclusion}
\label{sec:conclusion}

In this paper we have introduced a zero-shot system for 6D pose estimation aimed at tracking hands and spoons during real-world eating scenarios, which can assist in nutritional analysis. In doing so, we compare the performance of two SOTA video object segmentation models, Cutie and SAM2, and demonstrate that SAM2 produces both higher quality segmentations and more robust tracking when paired with BundleSDF. We also identify main sources of error within the system. We hope our contribution will lead to further developments for accessible and reliable nutritional tracking systems.
{
    \small
    \bibliographystyle{ieeenat_fullname}
    \bibliography{references}
}


\end{document}